\documentclass{article}
\usepackage{arxiv}
\usepackage[utf8]{inputenc} 
\usepackage[T1]{fontenc}    
\usepackage{hyperref}       
\usepackage{url}            
\usepackage{booktabs}       
\usepackage{amsfonts}       
\usepackage{nicefrac}       
\usepackage{microtype}      
\usepackage{xcolor}         
\usepackage{graphicx}
\usepackage{adjustbox}
\usepackage{caption}
\usepackage{subcaption}
\usepackage[numbers]{natbib}

\title{Gyan: An Explainable Neuro-Symbolic Language Model}

\author{%
  Venkat Srinivasan \\
  Innospark Ventures \& Gyan AI \\
  \texttt{venkat@innospark.vc} \\
  \And
  Vishaal Jatav \\
  Gyan AI Inc.\\
  \texttt{vishaal@gyanai.com} \\
  \And
  Anushka Chandrababu \\
  Gyan AI Inc.\\
  \texttt{anushka.chandrababu@gyanai.com} \\
  \And
  Geetika Sharma \\
  Gyan AI Inc. \\
  \texttt{geetika.sharma@gyanai.com} \\
}

\begin{document}

\maketitle

\begin{abstract}
  Transformer based pre-trained large language models have become ubiquitous.  There is increasing evidence to suggest that even with large scale pre-training, these models do not capture complete compositional context and certainly not, the full human analogous context.  Besides, by the very nature of the architecture, these models hallucinate, are difficult to maintain, are not easily interpretable and require enormous compute resources for training and inference.  Here, we describe Gyan, an explainable language model based on a novel non-transformer architecture, without any of these limitations.  Gyan achieves SOTA performance on 3 widely cited data sets and superior performance on two proprietary data sets.  The novel architecture decouples the language model from knowledge acquisition and representation.  The model draws on rhetorical structure theory, semantic role theory and knowledge-based computational linguistics. Gyan’s meaning representation structure captures the complete compositional context and attempts to mimic humans by expanding the context to a ‘world model’.  AI model adoption critically depends on trust and transparency especially in mission critical use cases.  Collectively, our results demonstrate that it is possible to create models which are trustable and reliable for mission critical tasks.  We believe our work has tremendous potential for guiding the development of transparent and trusted architectures for language models.
\end{abstract}

\section{Introduction}

Large language models based on a neural, generative architecture - BERT (\citep{devlin-etal-2019-bert}), ELMo \citep{peters-etal-2018-deep}, OpenAI/GPT  (\citep{radford_improving_2018}), Gopher (\citep{rae2022scalinglanguagemodelsmethods}), ERNIE (\citep{sun2021ernie30largescaleknowledge}), Llama (\citep{touvron2023llamaopenefficientfoundation}, \citep{touvron2023llama2openfoundation}), Gemma (\citep{gemmateam2024gemmaopenmodelsbased}), Qwen (\citep{bai2023qwentechnicalreport}), Mistral (\citep{Mistral}) have evolved rapidly reporting significant improvements in performance on a variety of NLP tasks (\citep{touvron2023llama2openfoundation}).  Despite these advances, there still remain significant limitations with such models, main among them being their lack of interpretability, explainability, hallucination, potential for misuse, inability to fully capture compositional ‘context’, and challenges relating to tractability (\citep{StochasticParrots}; \citep{GaryNonsense}; \citep{GaryDishonest}; \citep{SmithContextualWordReps}; \citep{DentellaInsensitivity}; \citep{Shani2025TradeCompression}).

More recently, LLMs have introduced Large Reasoning Models (LRMs) such as OpenAI’s o1/o3 (\citep{Jaech2026O1SystemCard}; \citep{OpenAI2024O1}), DeepSeek-R1 (\citep{Guo2025DeepseekR1}), Claude 3.7 Sonnet Thinking (\citep{Anthropic2025Claude37}), and Gemini Thinking (\citep{Google2025GeminiThinking}) with impressive results across reasoning benchmarks. However, the fundamental architectural characteristics and limitations of LRMs largely remain the same as that of the earlier versions of these models.

The implied hypothesis at the heart of these large models is that they can gain a human analogous level of knowledge and inferencing ability through sophisticated statistical models based on patterns and distributional semantics from a large amount of data (\citep{Mahowald2023DissociateLanguageandThought}, \citep{Gary2023MultimodalHallucinations}). \citet{brunila-laviolette-2022-company} observe that of the two approaches to distributional semantics - Harris’s approach (\citep{Harris1954}, \citep{Harris1988}) dominates the current makeup of LLMs.  Firth’s approach (\citep{Firth1957a}, \citep{Firth1957b} and \citep{Firth1957c}) is closer to models of human cognition where humans expand context both with episodic details and in an abstract fashion.

Human understanding of language critically depends on understanding the complete composition and the inter-related semantic roles of different constituent parts in the full composition.  In forming their understanding, humans also use their prior knowledge not explicitly stated in the document being processed.  Any machine meaning representation therefore has to be able to reflect the meaning of the full composition, incorporate relevant ‘global’ knowledge and allow for a level of abstraction.

In this paper, we describe Gyan, an explainable compositional language model, based on a novel, non-transformer architecture where the Gyan model is decoupled from data or knowledge. Gyan attempts to also explicitly construct a ‘world model’ in an attempt to reach human analogous understanding.  Gyan combines knowledge-based linguistics, thematic roles and rhetorical structure to create its rich and complete compositional encoding of meaning from natural language.

To demonstrate the \textbf{relative efficacy} of Gyan’s neuro-symbolic architecture, we provide comparative performance data from 3 widely cited - MS Marco (\citep{Bajaj2018MSMARCO}), PubMedQA (\citep{Jin2019PubMedQA}) and MMLU-Medicine (\citep{Hendryks2021MMLU}) data sets, a proprietary data set of 20 random queries and a real world essay grading data set.

Gyan-4.4 achieves SOTA performance on PubMedQA, MMLU-Medicine and is in the top 3 in MS Marco.  Gyan shows significantly improved relevance determination over search engines (re-ranking) on the 20 query data set.  Gyan also achieved 100\% acceptable results on a proprietary essay grading data set.

Collectively, these findings validate the efficacy of Gyan’s novel architecture for natural language in cases where explainability, tractability, transparency are critical.  The reasoning illustrated using these data sets demonstrate that the architecture can continually approach human analogous understanding of language as new knowledge is made accessible.

\section{Gyan LM decouples knowledge from its model}

Neural transformer-based LLMs are very large neural networks trained on an incredibly large amount of natural language content.  These LLMs are probabilistic models of word sequences and associations.  In this architecture, the model is inseparably coupled with data.  By pre-training these models on large amounts of data, these LLMs are presumed to have acquired generalizable ‘knowledge’ contained in the training data.

\begin{figure}
  \centering
  \includegraphics[width=\textwidth]{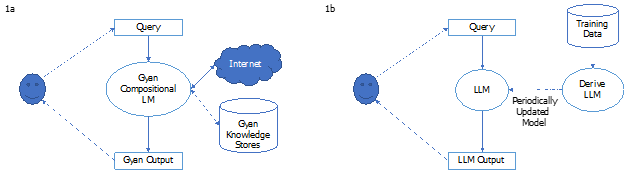}
  \caption{High Level Architecture of Gyan}
  \label{fig:gyan-hl-arch}
\end{figure}

In contrast, the Gyan model is not a statistical model of word patterns and is not pre-trained like the transformer based models are.  The Gyan architecture decouples the model from data (Figure \ref{fig:gyan-hl-arch}a).  The Gyan LM is a deep linguistic pipeline based on knowledge-based linguistics including syntactic representation, semantic roles and rhetorical structure.  The Gyan LM decomposes the entire document into a meaning representation graph (GMR) which preserves the composition fully.  With its deep linguistic pipeline, it can process any document in English (currently) without any training.  See the Methods section for a formal description of the model.

Gyan acquires its knowledge real time in three ways - (i) a base level of knowledge from dictionaries and ontologies; (ii) document collections which can be enterprise proprietary and/or public like PubMed, (iii) dynamically from the internet and other private sources accessible to the user.  In the case of the internet, Gyan LLM retrieves results using search engines and processes them real time.  All knowledge from pre-processed documents or acquired dynamically from the internet are stored in data stores referred to as 'Knowledge Stores' (KS), separate from the Gyan LLM.  These are transparent stores of GMRs of the documents processed using Gyan.

Pre-processed Knowledge Stores serve the same purpose in Gyan as pre-training in transformer LLMs but without the challenges associated with models based purely on word patterns.  Asymptotically, the amount of knowledge accessible to Gyan and neural LLMs will be the same if the training data used to train the neural LLMs are used to create equivalent Gyan 'knowledge stores'.

\begin{figure}
  \centering
  \includegraphics[width=\textwidth]{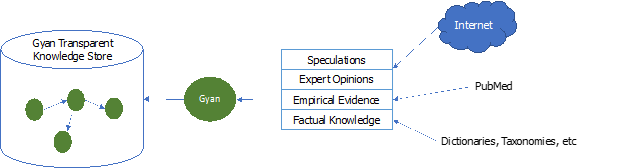}
  \caption{Knowledge Layers in Gyan}
  \label{fig:gyan-knowledge-layers}
\end{figure}

We can view Gyan’s knowledge repository in four layers as shown in Figure \ref{fig:gyan-knowledge-layers}.  The first layer is factual knowledge from dictionaries, taxonomies and other factual repositories.  The second level of knowledge is from expert studies most commonly in the form of academic research, e.g., PubMed.  The third layer of knowledge is expert opinions, e.g., analyst reports.  The final layer of knowledge can be speculative, e.g., blog posts and other social media content.  All four layers together form the knowledge network for that domain.

Gyan’s pre-processed knowledge stores enable it to provide fast response.  Gyan can behave as a small or large language model as a function of the number of KS that are used.

Because Gyan's meaning representation graphs are a deterministic representation of the underlying document and completely invertible, Gyan is fully explainable and traceable, will not hallucinate, cannot be manipulated or misused with data and is transferable across domains easily.  The Gyan LM needs a fraction of the compute resources neural LLMs need both for training and inference.

As is well known, updating transformer based models can cause instability (\citep{Kirkpatrick2017Catastrophic}).  In contrast, the Gyan LM is easily maintained.  With its decoupled and explicit meaning representation architecture, Gyan LM is able to reflect new knowledge continuously.  New knowledge can be maintained independent of existing knowledge or merged.

\section{Gyan LM attempts to mimic human knowledge organization and retrieval}

Human understanding of natural language discourse and organization of knowledge in it, is a combination of two essential elements – the discourse or document on the one hand and pre-existing human knowledge on the other.

Humans interpret or understand a document in the context of what we know already or our pre-existing cognitive perception (\citep{DiMaggio1997Culture}; \citep{Paller2002Experience}).  Concepts or areas in the discourse we have no prior knowledge of, constitute new knowledge which we attempt to integrate with our existing knowledge (\citep{Dudai2015ConsolidationMemory}).  Such integration involves creating a cognitive perception of the new knowledge.  Over time, this consolidation also semanticizes memories such that episodic details fade away and what remains is a more abstract version of the encoded memory (\citep{Dudai2015ConsolidationMemory}).

Complete machine understanding of a composition needs a holistic understanding of the ‘concepts and relationships among them expressed in the composition using formal linguistics – semantic structure of the composition made up of semantic roles of words, phrases, paragraphs, and rhetorical arguments used to express relationships among the ‘concepts’ in the composition, using rules of semantics (\citep{Portner2005Meaning}).  Any approach to understand it by breaking up the composition into words or phrases or sentences will be less than complete, e.g., \citet{SmithContextualWordReps}.

Neural LLMs have attempted to address this largely by innovating methods to ingest more context at the time of training the models (\citep{Hashimoto2016PhraseEmbeddings}; \citep{Iacobacci2016EmbeddingsWSD}; \citep{Mancini2017EmbeddingWordsAndSenses}; \citep{Peters2019KnowledgeEnhancedWordRepresentations}; \citep{Xiong2019pretrainedencyclopedia}; \citep{Wang2021Kepler}; \citep{Yu2014EmbeddingsWithSemanticKnowledge}).   \citet{Hermann2013SyntaxInEmbeddings} suggest the use of Combinatory Categorical Grammar (CCG) based autoencoders.  \citet{SmithContextualWordReps} reviews the use of contextual word representations but concludes that it is an open question how much of the work of understanding can be done at the level of words in context.  Besides, while useful, ingesting more context at a word or even sentence level, does not fully reflect full compositionality and does not address other fundamental limitations of the architecture such as transparency and hallucination.

On a broader level, how well do transformer-based LLMs mimic human understanding or organization of knowledge?  Knowledge organization in these models is largely influenced by distributional semantics.  \citet{brunila-laviolette-2022-company} observe that the main proponents of distributional semantics - \citet{Harris1954}, \citet{Harris1988}, \citet{Firth1957a}, \citet{Firth1957b} and \citet{Firth1957c} had two very different approaches to defining distributional semantics and Harris’s approach dominates the current model frameworks.  Yet, Firth’s approach is closer to models of human cognition where humans expand context both with episodic details and in an abstract fashion.

\citet{DentellaInsensitivity} point out that good performance in benchmark data sets reflect memorized specialized knowledge but not necessarily a solid understanding of language, such that LLMs may fail at comparatively easier tasks (\textit{Moravecs paradox}). Reasoning in machines using these LLMs is harder for simple, effortless tasks that human minds do easily.  Using questions on a set of 26,680 short texts featuring high frequency linguistic construction, they find that LLMs perform at chance accuracy and waver considerably in their answers. Based on their findings, they conclude that despite their usefulness in various tasks, current AI models fall short of understanding language in a way that matches humans, and that this may be a lack of an ability to regulate grammatical and semantic information.

Recently \citet{Shani2025TradeCompression} argue that humans organize knowledge into compact categories through semantic compression by mapping diverse instances to abstract representations while preserving meaning.  With a novel information-theoretic framework, they find that while LLMs form broad conceptual categories that align with human judgment, they struggle to capture the fine-grained semantic distinctions crucial for human understanding.

Recent body of evidence clearly suggests that current models do not adequately capture the full compositional context and certainly not the expanded human analogous context.  The Gyan LM and knowledge architecture attempt to reflect both aspects of human cognition mentioned earlier – the full compositional context of the document and the expanded human ‘world model’.  Its meaning representation model reflects the full compositional semantics of the document.  The Gyan LM then expands the query and the document semantically using its stored knowledge to approximate a ‘model of the world’ for the composition the way a human would see it.

In constructing the full compositional representation, Gyan also creates an extended pathway from the specific language used in the document to more abstract descriptions of the various discourse elements, e.g. if ‘parrot’ is a concept in the document, Gyan’s meaning representation will construct a dynamic pathway in representation from parrot to ‘bird’ and to ‘collective noun’.  Similarly the rhetorical phrase ‘results in’ is mapped by Gyan to a more rhetorical relation type ‘causal’.  Gyan’s meaning representation coupled with the expansion to integrate prior knowledge yields a meaning representation structure that is analogous to human understanding.  The hierarchical pathway from the surface level language (‘parrot’) to more abstract concepts like ‘bird’ and ‘collective noun’ also address the need for knowledge organization to be flexible enough to go from fine-grained expressions to abstract versions.  We believe the Gyan architecture allows a seamless bridging of expressive fidelity and semantic compression \citet{Shani2025TradeCompression}.

\section{Gyan LM creates an explainable meaning encoding of the complete composition}

Gyan LM’s meaning encoding referred to as Gyan Meaning Representation (GMR) is fully explainable as it preserves the full composition in place.  It has its roots in Rhetorical Structure Theory (\citep{Mann1987RST}, \citep{Taboada2006RST}; \citep{dohare2017textsummarizationusingabstract}; \citep{issa-etal-2018-abstract}), Abstract Meaning Representation (\citep{banarescu-etal-2013}) and Knowledge Graphs (\citep{Ehrlinger-2016-Definition-KG}; \citep{Pan2017ExploitingLinkedData}).

To construct GMR, the Gyan model decomposes the document step by step starting with Discourse Units (DUs) and ending with the concepts and phrases in every sentence.  Gyan detects DUs, identifies sentences that belong to each DU, and breaks down the sentences into concepts and relations.  In the Gyan LM, DUs can be nested within DUs and are modeled as structural or semantic.  Relations can be at several levels – DU level, inter-sentence and intra-sentence.  The set of sentences which belong to a DUs are judged by the Gyan LM to be closely related semantically to the subject of the DU.  At the sentence level, Gyan LM breaks down sentences into concepts and relations.  Similar to DUs, concepts may have sub-concepts.

The Gyan LM defines a document ($D$) to be a set of Ideas or Discourse Units ($DU$) on a subject.  The $DU$s in turn are expressed as a set of sentences ($S$) and each sentence is composed of linguistic elements including concepts, roles and rhetorical relations $(E, R)$ between them.  For example, a simple sentence can be thought of as having a subject and an object which are linked through a verb phrase reflecting a rhetorical relationship type.  Both concepts and rhetorical relations (E, R) in turn are mapped to a hierarchy which goes from being highly fine-grained or descriptive to higher levels of aggregation or abstraction, e.g., $parrot \rightarrow bird \rightarrow collective noun$.

\begin{figure}
  \centering
  \includegraphics[width=\textwidth]{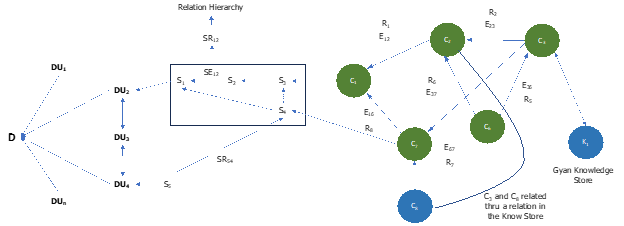}
  \caption{Gyan Meaning Representation Graph}
  \label{fig:gyan-gmr}
\end{figure}

Figure \ref{fig:gyan-gmr} shows an illustrative GMR.  From left to right, it shows the first level decomposition by Gyan of a document, D, into its discourse units, DUs.  DU2 shows the second level decomposition into a set of sentences which belong to the DU because of the strength of their semantic relationship to the DU’s head sentence.  SE12 is the surface level sentence relation between S1 and S2 which is mapped to an abstract relation type, SR12.  SR12 in turn maps to a hierarchy of increasingly more abstract relation types.

Next, Figure \ref{fig:gyan-gmr} illustrates the next level of decomposition where each of the sentences are decomposed into their constituent concepts and relations.

Gyan expands the sentences to reveal their compositional structure in the form of concepts and relationships between them.  Relationships between sentences are also effectively relationships between concepts in those sentences.  For example, a sentence starting with a pronoun refers to a person in the sentence it is related to.  A concept can be a noun or a phrase or a normalized form of a noun or verb phrase.

If a sentence can be mapped to multiple DUs, Gyan LM resolves the conflict by mapping to the DU with the higher semantic strength.  Gyan LM determines the strength of the semantic relationship using linguistic and rhetorical characteristics.  As an example, a sentence whose subject is a coreference of an object of a preceding sentence may be ranked lower in terms of relationship strength between those two sentences compared to a sentence whose subject is a direct coreference of the subject of the preceding sentence.  An important underpinning of the Gyan architecture is that word or phrase or sentence frequencies do not play any role.

The user can unpeel one section of the GMR at a time either based on concepts and/or rhetorical relationship types (R).  A multi-layered meaning representation allowing a gradual unpeeling of meaning also mimics human processing of natural language.  Often we read a document several times each time improving our understanding.

Figure \ref{fig:gyan-gmr} also illustrates how Gyan LM expands the meaning representation to create a ‘World Model’ for the document context.  Gyan LM automatically expands the GMR for both the user’s query and the documents being processed, using knowledge stores and content source(s) it has access to.  Gyan looks for related concepts outside of the current query or document which might have a bearing on the meaning representation for the current query or document.  This is illustrated for a document in Figure \ref{fig:gyan-gmr} by the blue nodes.  The expanded GMR allows Gyan to establish semantic relationships between concepts in the document which cannot be inferred from processing the document alone.  Humans reading the document would have readily made the same association with their prior knowledge.

Gyan LM can create a synthesized GMR representing multiple documents.  The GMR from each document is merged substituting semantic co-references at a \textbf{concept-sense level}.  The consolidated GMR is often not a simple union of constituent GMRs.  It is a \textbf{union} of expanded GMRs without loss of knowledge or surface form. The consolidated GMR can have more nodes and more edges than a simple union of two GMRs.  This is illustrated in the Appendix section.

\section{Results}

The Gyan GMR can be decoded to perform a wide range of NLU tasks - searches for relevant documents, topic modeling, tagging, summarization, synthesis of multiple documents, question-answering/inferencing, document analysis or review e.g., essay grading, legal document markup and document generation.  We report results from several experiments below including 3 widely cited benchmark data sets – MSMarco, PubMedQA and MMLU-Medicine.

\subsection{Gyan achieves Top 3 performance in a zero-shot test on MS MARCO}

\begin{figure}
  \centering
  \begin{minipage}{.5\textwidth}
    \centering
    \includegraphics[width=0.8\linewidth]{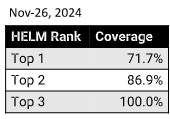}
    \captionof{figure}{Gyan 4.3 on MSMarco Passage Ranking}
    \label{fig:gyan-msmarco}
  \end{minipage}%
  \begin{minipage}{.5\textwidth}
    \centering
    \includegraphics[width=\linewidth]{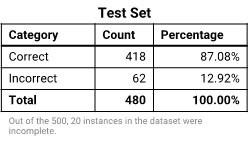}
    \captionof{figure}{Gyan 4.4 on PubMedQA dataset}
    \label{fig:gyan-pubmedqa}
  \end{minipage}
\end{figure}

To determine relative efficacy of the Gyan architecture, we benchmarked Gyan against the MS Marco dataset for passage ranking (\citep{Bajaj2018MSMARCO}) with minimal data in the Knowledge Store, without any additional domain specific knowledge and no training.  The only data in the Gyan Knowledge Store for this experiment were synonyms from WordNet (\citep{Miller1995WordNet}).  Gyan-4.3 was SOTA for 71.7\% of the queries, in the Top 2 on the leaderboard for 86.9\% of the queries and in the Top 3 for all the queries (Figure \ref{fig:gyan-msmarco}).

As Gyan Knowledge Stores grow in coverage for concepts in the MS Marco dataset, we can expect Gyan to further improve on the leaderboard.  The query-wise performance details are available from the authors on request.

\subsection{Gyan achieves SOTA performance on PubMedQA}

The PubmedQA dataset (\citep{Jin2019PubMedQA}) is a widely cided data set and contains all PubMed articles with question titles, and manually labeled answers for 1k of them for cross-validation and testing. In order to process the PubmedQA dataset, we created an initial Gyan Knowledge Store (KS) for medicine from 32 different domain specific datasets comprising medical dictionaries and MESH.  A detailed description of these data sets is available from the authors on request.

The PQA-L dataset of 1K questions and abstracts has been further divided into 500 for training and 500 for validation.  Gyan-4.4 used the KS for Medicine to answer the Test Set comprising 500 questions.  Since the Gyan architecture does not rely on pre-training, the pre-training data was not used.

As shown in Figure \ref{fig:gyan-pubmedqa}, Gyan-4.4 answered 87.08\% of the questions correctly.  As we can see from the table and from the leaderboard [https://pubmedqa.github.io/], as of Nov 30, 2024, Gyan-4.4 has already reached SOTA performance levels.  Gyan’s reasoning is transparent and understandable.  Illustrations of Gyan’s reasoning for a PubMedQA questions are available from the authors on request.

\subsection{Gyan achieves SOTA performance on MMLU-Medicine}

The Massive Multitask Language Understanding (MMLU) data set (\citep{Hendryks2021MMLU}) has gained significant popularity because it assesses the breadth and depth of language understanding capabilities across a diverse range of subjects.  For the purposes of this study, we only considered a subset comprising 9 clinical subjects as also done in \citet{Singhal-2025-MMLU-Medicine}.  The dataset contains 15908 questions in total, which are split into a few-shot development set, a validation set, and a test set. The few-shot development set has 5 questions per subject, the validation set is made of 1540 questions, and the test set has 14079 questions. Each subject contains 100 test examples at the minimum.  Each question contains 4 options out of which the model has to choose the correct one.

The nature of reasoning required for MMLU is different from PubMedQA.  MMLU requires the model to examine all options and rule in/out options.  PubMedQA required the model to provide a Boolean answer to the question based on the content in the associated abstract.

\begin{figure}
  \centering
  \includegraphics{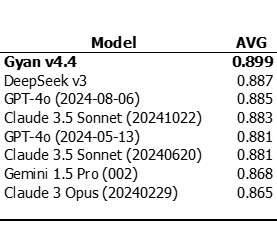}
  \caption{MMLU Leaderboard (May 12-2025)}
  \label{fig:gyan-mmlu-medicine}
\end{figure}

As shown in Figure \ref{fig:gyan-mmlu-medicine}, Gyan-4.4 achieves SOTA performance on the original MMLU data set.

Using the KS for Medicine, we evaluated Gyan LM on all the MMLU variants.  The detailed results are available from the authors on request.  Gyan-4.4 performance is stable across all these data sets validating the robustness of the Gyan architecture to errors in data.  Gyan-4.4 was relatively unaffected by purely representational edits like changes in the order of the options, increase in the number of options from 4 to 10 and also from errors in the data set.

Gyan-4.4 provides a detailed trace of its reasoning and inferencing for any question including in some cases declaring that there is not enough knowledge in the Knowledge Store to answer a question.  In such cases, Gyan also will suggest to the user that it can attempt to fetch the missing knowledge dynamically from the internet and other sources it may have access to.  A detailed trace for an example question is available from the authors on request.

\subsection{Gyan improves relevance ranking on a proprietary 20 Query Dataset}

Our hypothesis is that Gyan LM with its more complete meaning encoding will be better at finding relevant documents.  We tested this hypothesis on two data sets on a random set of 20 internet queries from various queries submitted by users of Gyan.  The MS MARCO passage ranking data set reported earlier is also a relevance classification data set.

\begin{figure}
  \centering
  \includegraphics[width=\textwidth]{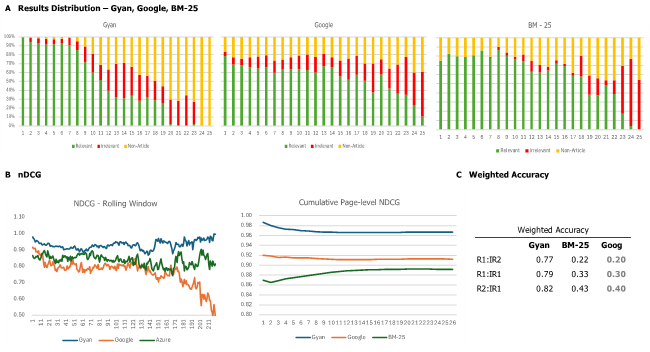}
  \caption{Ranking Results on the 20 Query Dataset [Google, MS Azure, Gyan]}
  \label{fig:gyan-m1-dataset}
\end{figure}

On the 20 random internet queries we used Google as the search engine to retrieve baseline results.  We selected queries with one, two and three words including queries with highly ambiguous words.  For each query, 2 human experts reviewed and labeled all the articles in the entire result set from the search engine(s).  Results from Google were compared with the labeled data set.  The data were then processed using the BM-25 re-ranker and compared with the labeled data.  Finally, the data were processed through Gyan.  [dataset available from the authors]

The page-wise distribution of results are displayed in Fig 9a.  The page here is the default organization priority in Google search.  In Figure \ref{fig:gyan-m1-dataset}a, green represents relevant articles, red irrelevant and yellow are non-articles.  The chart on the left shows the results according to Gyan, followed by Google and BM-25.  In the charts, we can see a very high portion of its results in earlier pages are relevant articles compared to BM-25 and Google.

We then computed the nDCG metric both on a cumulative basis and on a 20-item rolling window basis to remove any arbitrary pagination effect.  Windows of 10-items and 30-items did not show any material difference.  Figure \ref{fig:gyan-m1-dataset}b shows the nDCG metrics.  It shows the rolling nDCG and the cumulative nDCG by page.  We can see that the nDCG for Gyan outperforms both Google and BM-25.  Figure \ref{fig:gyan-m1-dataset}c displays the weighted accuracy metrics where we varied the relative costs for relevant and irrelevant classification from 1:2 for Relevant: Irrelevant, 1:1 and then 2:1. The metrics confirm that Gyan outperforms both Google and BM-25 on these queries across both the relevant and irrelevant categories.

\section{GMR provides traceable, explainable responses to queries}

Enterprise knowledge processed and stored in Gyan KS is a source of uncontaminated truth.  Gyan LM and KS enable an enterprise to create a responsible, trustable AI stack.

Gyan generates synthesized responses to queries, graphs for visual exploration, analysis of documents wrt a rubric, e.g., proposal evaluation, analytical summary of results as per a template, e.g., experiments, among other natural language tasks.  Synthesized responses to queries can be of different verbosity levels.  A comparative response from Gyan, GPT-4 and Gemini to the question ‘What is Brain Computer Interface’ is avaibale from the authors on request.  In the Gyan response, the trace back to the precise source document is illustrated only for the first sentence of the summary for brevity.  In addition to trace back to the exact locations from where the summary was created, Gyan can also provide complete reasoning for any part of its inference – why a specific source document was found relevant to the query, and how text from different documents were combined.

Gyan responses are extractive.  When multiple documents are combined, Gyan does so with very minimal paraphrasing to create a consolidated summary or synthesis.  Gyan of course removes redundant ideas in creating such a synthesis.  Extractive summaries are effectively a subset of the document text which summarize the full discourse.  The text is not modified in any way.

Gyan will not co-mingle responses from different KS unless needed to ensure that enterprise data is not co-mingled.

\section{Gyan requires negligible compute resources}

Most neural LLMs require an enormous amount of compute infrastructure for training and inference.  While the cost of training and inference is decreasing, it is still quite sizable.  The carbon impact of LLMs is a substantial issue. The surge in LLM adoption has, in turn, exacerbated the already considerable environmental impacts associated with machine learning (ML) (\citep{Thompson-2021-Diminishing-Returns}).

\begin{figure}
  \centering
  \includegraphics[width=\textwidth]{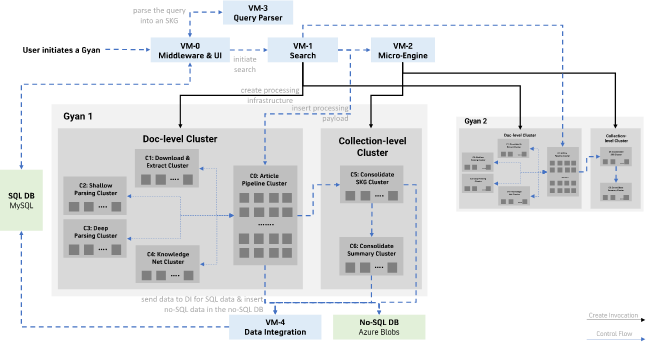}
  \caption{Gyan Physical Architecture}
  \label{fig:gyan-physical-architecture}
\end{figure}

Gyan, on the other hand, requires a negligible amount of compute resources both for training and inference.  Gyan does not need a GPU.  Gyan’s physical architecture is highly scalable, parallelized and the processing architecture auto scales with demand.  Gyan is thus highly energy efficient and sustainable from an environmental perspective.  Figure \ref{fig:gyan-physical-architecture} shows the details of the physical architecture.

\section{Discussion}

In this paper, we have presented a language model based on a novel architecture which addresses fundamental challenges with transformer-based language models, and attempts to create a closer human analogous machine meaning representation of knowledge. The architecture decouples the language model from knowledge and simulates a knowledge architecture that mimics humans.  Gyan moves away from purely statistical patterns of words to retain the complete composition in its surface form and expand it to a world model to fully understand the composition in a human analogous context.  It is closer to the Firth approach to distributional semantics rather than Harris.

We demonstrated the relative efficacy of the architecture by comparing Gyan performance on 3 widely cited benchmark data sets.  While there has been recent criticism of benchmarks as indicators of knowledge completeness, our primary goal is to demonstrate the relative efficacy of Gyan’s architecture. Gyan achieved SOTA performance in 2 of them and is in the Top 3 in the third.  The performance was achieved with limited amount of domain knowledge.  It is also important to note that Gyan only needs non-overlapping sets of such knowledge from credible sources contrasted with transformer based LLMs which need to be pre-trained on very large amounts of language data to be effective.  Gyan can continue to improve its performance on these data sets if it is provided the missing knowledge.  Missing knowledge is identified by Gyan explicitly in response to a user query.

We illustrated the transparency of every step in inference and reasoning in Gyan with examples from PubMedQA and MMLU-Medicine.  While the PubMedQA reasoning was to generate a Yes/No response, MMLU-Medicine required Gyan to evaluate each option and choose the best answer for the query.  We also demonstrated Gyan’s stability and invariance to perturbations in data representation e.g., changes in order or number of options, to errors in the data set.  The only aspect that will affects Gyan’s performance will be underlying errors in its knowledge store.

We also provided evidence of Gyan’s performance on a couple of real world proprietary data sets.  In a proprietary data set of 20 random search queries, Gyan meaning based relevance classification demonstrated significantly better results in finding more relevant documents compared to search engines.  We also found our model to be effective in analysis of documents with respect to a rubric in a small sample of open text responses in a higher educational setting.  These results are available from the authors on request.  The Gyan LM was able to correctly grade open text responses for 57 students in online courses across two globally respected higher educational institutions.  In one of these instances, Gyan LM grades were compared to 3 human graders and peer grades.  In the other case, Gyan LM grades were reviewed by the professor.

The ability of Gyan to function with very sparse data is a major advantage.  In a tactical sense, however this does require that Gyan locate and pre-process the knowledge that is relevant to an enterprise or use case. The completeness and accuracy of Gyan’s meaning representation is a function of how accurate the Gyan LM is in parsing the linguistic and rhetorical structure of documents.  While we have created an extensive set of pre and post processing rules to correct errors in syntactic parsing of language data, this will need to continue to improve over time.

Are there benefits for Gyan to operate in conjunction with transformer based LLMs?  We envision several scenarios for Gyan and transformer-based LLMs to co-exist.  First, Gyan KS can be used as the source of truth by an LLM.  Gyan KS can be integrated as part of an LLM as it is MCP compliant.  Gyan KS can be updated easily and remain completely transparent.  Based on how the LLM processes the Gyan KS, this can substantially reduce hallucination and also enable more precise traceability and explainability in the LLM output.  Second, Gyan LM can generate its normal output and also generate LLM output to look for any incremental knowledge that Gyan KS may not have obtained and attempt to validate such incremental knowledge.  It can include such additional knowledge in its output with an annotation on whether it was validated or not.  As a corollary, in a third scenario, Gyan LM can be used to validate and explain LLM output.  In this case, Gyan will process the LLM output through its normal process and attempt to validate as much of the LLM output as it can.  This can make some or all of the LLM output traceable and explainable.  The final scenario is for Gyan to create its meaning representation expansion of the user query to generate a contextually richer prompt to the LLM.

Does Gyan create bias?  Gyan does not generate bias in its processing.  But it does not eliminate bias that may be in the document being processed.  Since it is explainable and tractable, the user can immediately discern if the source data carries any bias.  Gyan also allows users to assign credibility levels to sources.

The Gyan LM is easily transferable across domains.  It will without any training be able to create a GMR of language data in a new domain, e.g., material science.  Gyan LM can incorporate domain specific reference data, domain specific rhetorical relationships which will make Gyan LM even more suitable for that domain.  Future work is being directed in two areas – building a large base level knowledge store which has factual knowledge on many concepts and in generating abstractive output similar to the transformer based LLMs but without losing the advantages of the novel Gyan architecture.

The Gyan LM is most appropriate for use cases where explainability, tractability, reliability, repeatability, privacy (no co-mingling) is mission critical.  The model architecture is a useful framework for not just language data but also for development of explainable and tractable models for other types of data.

\bibliographystyle{unsrtnat}
\bibliography{gyan}

\newpage

\appendix

\section{A Formal Description of the Gyan Language Model}

Gyan is a Hierarchical Hybrid Model. All levels of the hierarchy are Symbolic, except the lowest levels of hierarchy which may or may not rely on non-symbolic modeling. We describe the various components of the Gyan model in this method.

\section{Knowledge Representation}

\subsection{Discourse Models}

\subsubsection{\textbf{Abstract Discourse Models}}

A Discourse Model is a blueprint used to define a specific class of documents, such as web documents, text books,  legal contracts, legislation, scientific papers, resumes, etc.  It contains the specifications to identify, parse and understand the structure and relationships between the structures, within any document that belongs to this Discourse Model.  It is defined as the following 3-tuple:

\begin{center}
 $DM=(DModel_{ID,\ DM},{\widehat{DU}}_{DM},{\widehat{DE}}_{DM})$
\end{center}

\begin{quote}
 where:
 \begin{itemize}
  \item
  	$DModel_{ID}$ are the identification rules
  \item
	${\widehat{DU}}_{DM}$ is the collection and definition of all the various discourse units in $DM$.
  \item
	${\widehat{DE}}_{DM}$ is the collection of all the properties and relationships between the discourse units in $DM$.

 \end{itemize}

\end{quote}

The $DModel_{ID,DM}=\{id_1,id_2,\ldots,id_n\}$ is a collection of rules $id_i$ that determine whether a given document belongs to the Discourse Model $DM$.  Each $id_i\in DModel_{ID},id_i:F\rightarrow IDO$ can be modelled as a function that takes a file $f\ \in F$ and determines whether the document in the file belongs to $DM$ through $IDO=\{ido_1,ido_2,\ldots,id_m\}$.

Each $ido_i\in IDO$ is a tuple $ido_i=(range_i,tf_i)$ where a $range_i$ is a representation of a range of offsets within the file $f$ and $tf_i\in\{true,false\}$ is a flag on whether the contents of the file $f$ within the range $range_i$ belongs to the $DM$ through a Boolean value.

The ${\widehat{DU}}_{DM}$ defines the set of fundamental and semantic block of discourse within $DM$.  Each ${\widehat{du}}_i\in{\widehat{DU}}_{DM}$ is a 3-tuple of the form ${\widehat{du}}_i=(NAME,ATTR_i,DETECT_i,ID_i)$ where $NAME$ is the human-readable name of the ${\widehat{du}}_i$, $ATTR_i=\{a_1,a_2,\ldots,a_\gamma\}$ are a set of definitions of attributes associated with a discourse unit, $DETECT_i=\{dt_1,dt_2,\ldots,dt_k\}$ are a set of rules to detect the boundaries of each instance of ${\widehat{du}}_i in f$ and $ID_i=\{idf_1,idf_2,\ldots,idf_\alpha\}$ are a set of rules to identify the $identifier$ (e.g. clause 3.1(a), etc.) of each of the instance of ${\widehat{du}}_i in f$).  A discourse unit definition $du_i\in{\widehat{DU}}_{DM}$ with the $NAME=\mathrm{Name}$  can also be denoted as $Name_{du}$.

The ${\widehat{DE}}_{DM}$ defines the set of different types of relationships between discourse units for this $DM$.  Each ${\widehat{de}}_i\in DM$ is a 4-tuple of the form ${\widehat{de}}_i=(NAME,REL_{DM},CAL_{DM},R_{DM})$, where $NAME$ is the human-readable name of the ${\widehat{de}}_i, REL_{DM}$ is the set of all the relationships that can exist between two discourse units $du_i,du_j\in{\widehat{DU}}_{DM}$, $CAL_{DM}$ is the calculus on all the relations $rel_i\in REL_{DM}$ and $R_{DM}$ are the set of rules to detect the instances of relations between two instances of discourse units $du_i,du_j\in{\widehat{DU}}_{DM}$ within a document.  The $CAL_{DM}=\{cal_1,cal_2,\ldots,cal_\beta\}$ are a list of first-order logic rules on relations and discourse units.  A discourse edge definition $de_i\in{\widehat{DE}}_{DM}$ with the $NAME=\mathrm{Name}$ can also be denoted as $Name_{DE}$.

\subsubsection{\textbf{Default Textual Discourse Model}}

The Default Text Discourse Model $DM_{DT}$ is the most generic discourse model for text documents.  Several, more complex Textual Discourse Models can be transformed into $DM_{DT}$ by mapping the discourse units and discourse units and discourse relations into those of $DM_{DT}$.

The discourse units definitions ${\widehat{DU}}_{DT} \in DM_{DT}$ contains definitions of discourse units like $\{$ $Document_{DU}$, $Section_{DU}$, $TopicCluster_{DU}$, $Enumeration_{DU}$, $Example_{DU}$, $Table_{DU}$, $Figure_{DU}$, $\ldots\}$.  The discourse edge definitions ${\widehat{DE}}_{DT}={\widehat{DE}}_{DT_{Struct}}\cup{\widehat{DE}}_{DT_{Disc}}$ are made up of two distinct sets of discourse edge definitions.  The ${\widehat{DE}}_{DT_{Struct}}$ contains information about the structural properties of discourse units or structural relations between them, e.g. ${\widehat{DE}}_{DT_{Struct}}=\{ Contains_{DE}, Preceeds_{DE}, \ldots\}$.  The ${\widehat{DE}}_{DT_{Disc}}$ contains information about discourse properties or relationships between discourse units, like those from the Rhetorical Structure Theory, e.g. ${\widehat{DE}}_{DT_{Disc}}=\{$ $Elaboration_{DE}$, $Justification_{DE}$, $Background_{DE}$, $Evidence_{DE}$, $Summary_{DE}$, $\ldots\}$.

\subsubsection{\textbf{Examples of other Textual Discourse Models}}

The Abstract Discourse Model DM harps on the fact that a document written with a particular type of intent or to serve a particular function will always have the same elements of discourse in them.  These elements can be explicit or implicit.  The DM can be instantiated to describe a wide variety of textual documents, including the most niche ones.  Following are some examples:

\begin{itemize}
  \item
  \textit{Web Document Discourse Model} $DM_{Web}$ will hold information on discourse units like $Header_{DU}$, $Footer_{DU}$, $NavigationalArea_{DU}$, $Advertisements_{DU}$, $MainArticle_{DU}$, etc. and discourse edges ${\widehat{DE}}_{Web}$ like $LeftOf_{DE}$, $TopOf_{DE}$, $DocumentLink_{DE}$, etc.
  \item
  \textit{Research Paper Discourse Model} $DM_{Research}$ will have units like $Abstract_{DU}$, $Conclusion_{DE}$, $Methods_{DU}$, $PriorWork_{DU}$ and edges like $Citation_{DE}$, $BibLink_{DE}$, etc.
  \item
  \textit{Legal Contract Discourse Model} $DM_{LegalContracts}$ will have units like $Clause_{DU}$, $Definitions_{DU}$, $Appendix_{DU}$ and edges like $Ammends_{DE}$ and $Cites_{DE}$.
  \item
  Other examples include one for each of \textit{Resumes}, \textit{Legislations}, \textit{Regulations}, \textit{User Manuals}, \textit{Patent Documents}, every \textit{Tax Form}, etc.
\end{itemize}

\subsubsection{\textbf{Multimodality in Abstract Discourse Models}}

The Discourse Model abstraction unifies documents with different underlying formats, such as a relational database (structured) and a text report (unstructured).  This is achieved by defining a model $DM$ whose universe of units, ${\widehat{DU}}_{DM}$, contains specifications for discourse units from both sources, e.g. a $PatientSafetyReport_{DU}$ row from a FDA Database and $Document_{DU}$ from the text report.

The edge detection rules $R_{DM}$ in the discourse edge definitions ${\widehat{DE}}_{DM}$ could determine a specific relation from a row (a particular instance of $PatientSafetyReport_{DU}$), a column (an attribute ATTR in all the instances of $PatientSafetyReport_{DU}$) or a cell, to a part of the text report.

The Discourse Model abstraction also generalizes to all modes of data - images, mathematical expressions, audio and other modes of data.  The instantiation of the Discourse Models will have mode specific characterization, rules and implementations.

\subsection{Knowledge}

\subsubsection{\textbf{Abstract Knowledge Bases}}

An \textbf{Abstract Knowledge Base} contains a schema of storing Knowledge in a structured form and is defined as:

\begin{center}
 $ KB = \{N,R,X,CAL,KG,D_{Collection},Exp_{Collection}\} $
\end{center}

\begin{quote}
 where:
 \begin{itemize}
  \item  $N$ is the universe of all the normalized concepts,
  \item  $R$ or the universe of all the possible relations,
  \item  $X$ or the universe of all the contextual frames and
  \item  $CAL$ or the calculus that contains all the relations between relations and concepts.
 \end{itemize}

\end{quote}

The $N$ or the Normalized Concepts Universe contains all the specifications of the concept it represents.  Every concept $n_i\in N$ is a tuple $n_i=(spec_{text},spec_{audio},spec_{visual},n_{hypernym},attr_i)$.  The first three elements contain textual, audio and visual specifications of $n_i$.  These multi-modal specifications of the concept contain enough necessary information to correctly identify the concept from any instance of a discourse unit in any document.  The $n_{hypernym}$ contains references to all other $n_j\in N:n_j\preceq_{hypernym}n_i$, thereby making $N$ a poset or a partially ordered set.  The ${attr}_i=\{kv_1,kv_2,\ldots,kv_\epsilon\}$ is the set of all key-value pairs defining other constraints or properties of this concept.

The $R$ or the universe of all possible relations contains all relations including semantic, episodic and procedural relations.  Every $r_i\in R$ is a tuple $r_i=(spec,r_{hypernym},attr_i)$.  The first element contains all the specs and information to detect the instance of the relation unambiguously.  The $r_{hypernym}$ contains the references to all other $r_j\in R:r_j\preceq_{hypernym}r_i$ such that $r_j$ is either a hypernym of $r_i$ or is equivalent to it, thereby making $R$ a partially ordered set.  The ${attr}_i=\{{kv}_1,{kv}_2,\ldots{kv}_\zeta\}$ is the set of all key-value pairs defining other constraints or properties of this relation.

The $X$ is the universe of all the possible semantic frames or dimensions, that provides completeness to a relation between two concepts.  $X=\{F_{Temporal},F_{Spatial},F_{Modality},F_{Deontic},F_{Aspectual},\ldots\}$ is made up of multiple primitive frames, that can be combined to create a complete Contextual Frame $cx=\{f_1,f_2,\ldots,f_\eta\}$ where each $f_i\in cx$ is a key-value pair of the primitive frame name and its value.

The $CAL$ is the relational calculus that enables reasoning and provides a means to generate new knowledge from existing ones.  $CAL=(AX,\ IR)$.  $AX$ or \textit{axioms} are first-order logic statements about relations in $R$.  $IR$ are the inference rules that allow for deductions on relations in $R$ and on contextual frames.  Every inference rule $ir\in InferenceRules$ is a tuple of $ir_i=(P,C)$ where $P=\{p_1,p_2,\ldots,p_\theta\}$ are a series of first-order logic premises and $C=\{c_1,c_2,\ldots,c_\kappa\}$ are a series of first-order logic conclusions.

The $KG=(N_{KG},E_{KG})$ is the knowledge graph where the nodes are $N_{KG}\subset N$ and the edges are $E_{KG}$.  The edges $e_{i,j}\in E_{KG}$ is the edge between $n_i,n_j\in N_{KG}$ and is represented by the tuple $e_{i,j}=(n_i,n_j,r_{i,j},\tau_{i,j})$, where $\forall e_{i,j}\in E_{KG},r_{i,j}\in R$.  An additional information with every relationship is the trace $\tau_{i,j}$, that has the full physical trace of the source of this relation.  This trace can be of two types:  from a document or from an expert.  Additionally, the relation could be coming from several documents and several experts as well.  Therefore, every $\tau_k=\{\tau_{k_1},\tau_{k_2},\ldots,\tau_{k_\epsilon}\}$.  The references to documents are stored within the $D_{Collection}$ and references to experts are stored within ${Exp}_{Collection}$.

In case of a document trace, every $\tau_{k_m}$ is of the form $\{f_m,v_{f_m}\}$, where $f_m$ is the reference to the physical file of the trace document and $v_{f_m}$ is the exact logical document block within the trace document $f_m$ from where the relationship was either inferred or selected.  All such traced documents are stored within the $D_{Collection}$.  In case of an expert annotation, every $\tau_{k_m}$ is of the form $\{{exp}_m,comments\}$, where ${exp}_m$ is a reference to the expert and comments contains some rational behind the addition of this relation into the Semantic Knowledge Net ${KN}_{sem}$.

Multiple instances of simpler Knowledge Base can come together to create a composite Knowledge Base instance.  Every instance of a Knowledge Base is represented as $kb_i=\{N_i,R_i,X_i,CAL_i,KG_i,D_{Collection,i}\}$.  Given two knowledge bases $kb_i$ and $kb_j$, a composite knowledge base $kb_{i,j}=\{N_{i,j},R_{i,j},X_{i,j},CAL_{i,j},KG_{i,j},D_{Collection,i,j}\}$ is formed by taking a union of the corresponding constituents, while consolidating semantically equivalent entries, defined by the operator $U_{Sem}$.  Therefore, $N_{i,j}=N_i\cup_{Sem}N_j$, $R_{i,j}=R_i\cup_{Sem}R_j$, $X_{i,j}=X_i\cup_{Sem}X_j$, $CAL_{i,j}=CAL_i\cup_{Sem}CAL_j$, $KG_{i,j}=(N_{KG,i}\cup_{Sem}N_{KG,j},E_{KG,i}\cup_{Sem}E_{KG,j})$,  $D_{Collection,i,j}=D_{Collection,i}\cup_{Sem}D_{Collection,j}$ and $EXP_{Collection,i,j}=EXPCollection,i\cup_{Sem}EXP_{Collection,j}$.

\subsubsection{\textbf{Universal Knowledge Net}}

Universal Knowledge Net $(KN_W)$ is the repository of all the possible knowledge in the world, including the knowledge of all the discourse models.  It comprises of $KN_W=\{$ $N_W$, $R_W$, $X_W$, $CAL_W$, ${KN}_{Sem}$, ${KN}_{Epi}$, ${KN}_{Proc}$, $DM_W$, $D_{Collection,\ W}$, $EXP_{Collection,\ W}$ $\}$, where $N_W$ is the collection of instances of all the concepts in the world, $R_W$ is the collection of instances of all the relationships in the world, $X_W$ is the collection of all the semantic frames in the world and $CAL_W=(AX_W,IR_W)$ is the calculus containing instances of all the axioms and inference rules in the world on $N_W$, $R_W$ and $X_W$.

The ${KN}_{Sem}$ or the \textit{Semantic Knowledge Net} models static, factual knowledge about concepts and their relationships, independent of time or specific events.  The Semantic Knowledge Net is part of the Universal Knowledge Net and is defined by ${KN}_{Sem}=($ $N_{Sem}$, $R_{Sem}$, $X_{Sem}$, $CAL_{Sem}$, ${KG}_{Sem}$, $D_{Collection,Sem}$, $EXP_{Collection,Sem})$, where  $N_{Sem}\subset_{Sem}N_W$, $R_{Sem}\subset_{Sem}R_W$, $X_{Sem}\subset_{Sem}X_W$, $CAL_{Sem}\subset_{Sem}CAL_W$, $D_{Collection,Sem}\subset_{Sem}D_{Collection,W}$ and $EXP_{Collection,Sem}\subset_{Sem}EXP_{Collection,W}$.  Note that since $K_{Sem}$ is a repository of only generalized, factual and foundational knowledge, the $X_{Sem}=\ \emptyset$.

The $KN_{Epi}$ or the \textit{Episodic Knowledge Net} models episodic knowledge of the world, such as events, states, etc.., and is described as the knowledge base ${KN}_{Epi}=(N_{Epi},R_{Epi},X_{Epi},CAL_{Epi},{KG}_{Epi},D_{Collection,Epi},EXP_{Collection,Epi})$.  All the properties of $KN_{Epi}$ are same as $KN_{Sem}$, except that the $X_{Epi}\neq\ \emptyset$ .

The $KN_{Proc}$ or the \textit{Procedural Knowledge Net} models knowledge about real world processes, instructions, workflows, etc. and is described as the knowledge base ${KN}_{Proc}=(N_{Proc},R_{Proc},X_{Proc},CAL_{Proc},{KG}_{Proc},D_{Collection,Proc},EXP_{Collection,Proc})$.  All the properties of $KN_{Proc}$ are same as $KN_{Sem}$, except that the $X_{Proc}\neq\ \emptyset$  and the edges $e_{i,j}$ in ${KG}_{Proc}$ contain an additional property called \textit{procedural context}.  The edges $e_{i,j}\in E_{Proc}$ in ${KG}_{Proc}$ of $KN_{Proc}$ are edges between $n_i,n_j\in N_W$ and are represented by the $tuple e_{i,j}=\{n_i,n_j,r_{i,j},\tau_{i,j},px_{i,j}\}$.  Some peculiar relations in the ${KN}_{Proc}$ include the ones like $has_step$, $requires_input$, $is_precondition_for$, etc.  The representation includes an additional property called \textit{procedural context} or:

\begin{center}
  $ px_{i,j}=\{fr_{i,j},seq_{i,j},conditionality_{i,j},precon_{i,j},postcon_{i,j}\} $
\end{center}

\begin{quote}
 where:
 \begin{itemize}
  \item $fr_{i,j}$ defines the contextual frames (as before),
  \item $conditionality_{i,j}$ has details about logical branches like $if_else$, etc.,
  \item $precon_{i,j}$ defines preconditions before a step of the procedure is to be executed and
  \item $postcon_{i,j}$ defines postconditions after the step of the procedure is executed.
 \end{itemize}

\end{quote}

The $DM_W$ or the \textit{Discourse Models Universe} is the knowledge of all the discourse models in the universe.  The $DM_W=\{DM_W,DME_W\}$ is a graph where the nodes are the universe of Discourse Models $DM_W=\{{dm}_1,{dm}_2,\ldots,{dm}_\delta\}$ and the edges are relations from ${DME}_W$ between the discourse models themselves.  The ${DM}_W$ models relationships between various discourse models, such as, ${DM}_{LegalContract}\ -----\ specializes\dashrightarrow DM_{Text}$ and $DM_{Web}-----\ links_{to}\dashrightarrow DM_{video}$.  This graph structure allows for unifying inferences across document types and content present in different file-formats.

\subsubsection{\textbf{Null Knowledge Net}}

A \textbf{Null Knowledge Net} or $KN_{\phi,\ B}=\{N_B$, $R_B$, $X_B$, $CAL_B$, $KN_{Sem,\phi}$, $KN_{Epi,\phi}$, $KN_{Proc,\mathrm{\Phi}}$, $DM_{Min}$,$D_{Collection,\mathrm{\Phi}}$, $EXP_{Collection,\mathrm{\Phi}}\}$ is a state of a Knowledge Net when no knowledge has been added into the Knowledge Net.

It contains an empty knowledge net, that contains a default set of base universal concepts $N_B$, relations $R_B$ and frames $(X_B)$.  It also contains no semantic $(KN_{Sem,\ \mathrm{\Phi}})$, episodic $(KN_{Epi,\Phi})$ or procedural knowledge $(KN_{Proc,\Phi})$ as well.  And therefore, it contains an empty document collection $(D_{Collection,\Phi})$ and the expert collection $(EXP_{Collection,\Phi})$.  However, it contains a minimal universe of discourse models $(DM_{Min})$, which is format-specific default discourse models.

The base concepts, relations and frames are a set of concepts and relations that has been identified by an implementation.  Some examples of format-specific default discourse models include a collection of Textual Discourse Models for HTML, PDF, Imaged Document Discourse Model or Open Office formats, a minimal Image Discourse Model and a minimal Video Discourse Model.  This Null Knowledge Net forms the basis of bootstrapping and iterative building of an instance of Knowledge Net from several documents.

\subsection{Document}

\subsubsection{\textbf{Abstract Document Graph}}

The \textbf{Abstract Document Graph} or $AG_{Doc}$ is a multi-layered graph, which is computed in the context of the provided Knowledge Net or $KN$.

\begin{center}
  $ AG_{Doc,KN} = (F,DM_{Doc},DG,C_{Doc},{ED}_{Doc}) $
\end{center}

\begin{quote}
  where:
    \begin{itemize}
      \item  $F$ is the physical file that contains the document,
      \item  $DM_{Doc}$ is the Discourse Model from KN’s DM database to which Doc belongs, and
      \item  $DG$ is the Discourse Graph of the Doc as constructed based on the $DM_{Doc}$ specifications.
    \end{itemize}
\end{quote}

A \textbf{Gyan Meaning Representation Graph} for a document or \textbf{Document GMR} or $Doc_{KN}$ of a Document $d$ is the instance of the Graph of the document, grounded on the knowledge net $KN$.  Each file $f \in F$ is a physical file that is broken down into a set of primitive graphical elements $\{e_1,e_2,\ldots,e_\alpha\}$.  These can represent different things based on the format of a file.  For example, for a PDF file, these could represent instances of text blocks, SVG graphics, etc.; for an image, this could represent blobs of objects in the form of pixel hulls, etc.

The $DM \in KN$ contains a collection of all the discourse models.  Each $dm \in DM$ in turn contains a list of $id_i$ which are rules to identify whether a document file $f_j$ belongs to dm.  The Abstract Document Graph contains the reference of $DM_{Doc}$ to which it belongs to, such that all the identification rules $id_j \in DModel_{ID,DM}$ are satisfied by the file and the document.  This discourse model comes with a set of rules to identify discourse units and discourse edges through ${\widehat{DU}}_{DM}$ and ${\widehat{DE}}_{DM}$ respectively.

The $DG = (DU, DE)$ or the Discourse Graph of the file $f$ is the first layer of the Document Graph, which is defined by the graph with $DU$ as nodes and $DE$ as edges.  All the $du_i \in DU$ and $de_{i,j} \in DE$ conform to types as specified in ${\widehat{DU}}_{DM}$ and ${\widehat{DE}}_{DM}$ respectively.

Every $du_i \in DM$ is a tuple $du_i = ({NAME}_i, {ATTR}_i, CG_i)$ where ${NAME}_i$ is the name of the type of the discourse unit, ${ATTR}_i = \{{kv}_1, {kv}_2, \ldots, {kv}_\beta\}$ are a collection of key-value pairs that provide values to various attributes of this discourse unit and $CG_i$ is the concept graph of all the concepts contained within $du_i$.  Similarly, every $de_{i,j} \in DE$ is a tuple $de_{i,j} = \{du_i, du_j, NAME_{DM_{Doc},i,j}, r_{DM_{Doc}, i, j}, ATTR_{i,j}, x_{ij}, x_{ij,norm}\}$ where the $du_i$ and $du_j$ are the discourse units within the document, $NAME_{DM_{Doc},i,j}$ is the name of the discourse relation between $du_i$ and $du_j$, $r_{DM_{Doc},i,j}$ is the type of relationship as defined in ${\widehat{DE}}_{DM}$, $ATTR_{i,j} = \{kv_1, kv_2, \ldots, kv_\epsilon\}$ is a set of key-value pairs for that provide values to the various attributes of this discourse edge and $x_{ij}$ and $x_{ij,norm}$ are the contextual frames of the relation between the discourse units.

The ${CG}_i = ({CON}_i, {ED}_i)$, or the Concept Graph of a Discourse Unit, is the second layer of the Document Graph with ${CON}_i \subset C_{Doc}$ as nodes and ${ED}_i$ as edges.  The scope of the ${CG}_i$ is the ${du}_i$.  The ${CON}_i = \{c_{i1}, c_{i2}, \ldots, c_{im}\}$ is a set of all the concepts that occur within the ${du}_i$ and ${ED}_i = \{{ed}_{i1}, {ed}_{i2}, \ldots, {ed}_{ik}\}$ is a set of all the edges between concepts of ${du}_i$.  Note that $\forall du_i \in DU, C_{Doc} = \cup{CON}_i and ED_{Doc}=\cup ED_i$.

Every concept $c_{ij} \in {CON}_i$, or a concept occurring within $du_i \in DU$, is defined by the tuple $c_{ij} = (c_{name}, c_{norm}, c_{attr}, c_{trace})$ where $c_{name}$ is the literal concept used within $du_i$, $c_{norm}$ is the reference to a normalized concept $norm \in N$ from the provided knowledge net $KN$, $c_{attr} = \{kv_1, kv_2, \ldots, kv_\gamma\}$ is a set of key-value pairs of attributes and values and $c_{trace} = \{e_1,e_2,\ldots,e_\delta\}$ is the reference to the primitive graphical elements in the file $f$ containing the physical document, where the concept occurred.

Every edge ${ed}_{i,x,y} \in {ED}_i$, or an edge between $c_{ix} \in {CON}_i$ and $c_{iy} \in {CON}_i$, is defined by the 5-tuple ${ed}_{i,x,y} = \{c_{ix}, c_{iy}, r_{i,norm}, x_i, x_{i,norm}, type\}$, where $r_{i,norm} \in R \in KN$ is the reference to one of the sets of relations defined within the $KN$, $x_i$ is the contextual frame of this relationship, $x_{i,norm} \in X \in KN$ is the reference to one of the sets of contextual frames within the $KN$ and $type=\{Unknown, Semantic, Episodic, Procedural\}$ is the type of information that this edge represents.

There could be a concept $c_{ij}$ whose $c_{norm} = \phi$ or $null$ implying that it couldn’t be mapped to any of the normalized concepts.  Similarly, there could be an edge $r_{i,x,y}$ whose $r_{i,norm}= \phi$ and/or $x_{i,norm} = \phi$ and/or $type = \phi$, implying that the relation, frame and/or type couldn’t be normalized.  In cases where a concept, relation or a frame is not normalized to the ones in the provided knowledge net, they are treated as localized and serves as an indication of incomplete context or knowledge within the $KN$.

\subsubsubsection{\textbf{\textit{Contextual Dimensions of Document GMR}}}: The frame and type can be considered as two dimensions based on which the multi-layered document semantic knowledge graph can be segmented into multiple graphs.  Examples of segments can be past tense episodic document graph, semantic document graph, obligations document graph, etc.  Each of these slices could serve explicit purpose for a particular use-case of inferencing or knowledge acquisition.  Each dimension can be represented as $Doc_{KN,x,type}$.

\subsubsubsection{\textbf{\textit{Conceptual Abstraction of a Document GMR}}}: A document’s semantic knowledge graph can be transformed into a conceptual abstraction by replacing all the concept nodes and relations with their hypernym.  Note that all the concepts $C_{Doc}$ and relations within $ED_{Doc}$ within the document belong to $N \in KN$ and $R \in KN$ and both $N$ and $R$ are ordered sets based on the relationship of hypernym.  The space of all the conceptual abstractions of a document SKG is defined by $CA_{D_{KN}}$ and $Doc_{KN} \in CA_{D_{KN}}$.  This conceptual abstraction becomes a key tool to perform analyses across multiple documents.

\subsubsection{\textbf{Knowledge Net from a Collection of Documents}}

A \textbf{Single Document Base Knowledge Net} or \textbf{$KN_{d,B}$} is a Knowledge Net that is seeded with knowledge from a single, specific document.  The $KN_{d,B}=\{N_B, R_B, X_B, CALC_d, KN_{Sem,d}, KN_{Epi,d}, KN_{Proc,d}, DM_d, \{d\}, EXP_{Collection,\mathrm{\Phi}}\}$ is formed by taking the Document Graph of the Document ${DG}_{d,\mathrm{\Phi}} = (F, {DM}_d, DG, C_d, {ED}_d)$ and identifying generalizable semantic, episodic and procedural knowledge from d and populating it into a $KN_{\Phi,B}$.

A \textbf{Document Collection Base Knowledge Net} or \textbf{$KN_{D_C}$} is a knowledge net that is formed by populating a $KN_{\Phi,B}$ with generalizable semantic, episodic and procedural knowledge from all the documents from a given collection of documents $D_C = \{d_1, d_2, \ldots, d_\zeta\}$.

\subsubsection{\textbf{Knowledge Store of a Collection of Documents}}

\textbf{Base Knowledge Net} or \textbf{$KB_{D_C}$} whole Document Collection $D_{Collection, D_C}$ contains the Document’s individual Semantic Knowledge Graphs.  Therefore, if $D_C = \{d_1, d_2, \ldots, d_\eta\}$, then $D_{Collection,D_C} = \forall d_i \in D_C, \cup{Doc}_{d_i}$.  The difference between \textbf{Document Collection’s Base Knowledge Net} and \textbf{Document Collection’s Knowledge Store} is the presence of and access to individual document’s semantic knowledge graphs.  Therefore, $KN_{D_C} = KN_{D_C} \cup_{collection} \{\forall d_i \in D_C, Doc_{d_i}\}$.

\section{Gyan Pipeline for Textual Content}

\subsection{Meaning Graph Encoder}

Gyan’s \textbf{Meaning Graph Encoder} (\textbf{$Encoder$}) for Textual Content takes a Knowledge Net ($KN$) and a physical file ($F$) and generates a Document Semantic Graph ($Doc_{KN}$) in the context of $KN$.

\begin{center}
	$Encoder : (F \times KN) \rightarrow Doc_{KN}$
\end{center}

The $Encoder$ is a sequential pipeline of multiple steps, where the output of every step serves as the input of the next one.  The whole process is summarized in Table 1.

\begin{table}
  \centering
  \begin{adjustbox}{width=\textwidth}
    \begin{tabular}{|c|}
      \hline
      \\ [2.5pt]
      $ {Doc}_{KN}=(F,{DM}_F,DG,C_{Doc},{ED}_{Doc})$ \\ [2.5pt]
      $ KN=(N,R,X,CAL,{KN}_{Sem},{KN}_{Epi},{KN}_{Proc},DM,D_C,{EXP}_C)$ \\ [2.5pt]
      $ Encoder:F\times KN\rightarrow Doc_{KN}$ \\ [2.5pt]
      $ Encoder=DMID\ \circ Extr\ \circ LingSimp\ \circ ConceptDetect\ \circ DGGen\ \circ CGGen\ \circ CFraming\ \circ SemGrounding$ \\ [2.5pt]
      $ DMID:F\ \times KN\rightarrow DM_F$ \\ [2.5pt]
      $ Extr:F\times DM_F\rightarrow H$ \\ [2.5pt]
      $ LingSimp:\ H\ \rightarrow\ U$ \\ [2.5pt]
      $ ConceptDetect:U\rightarrow V\times C_{Doc}\times RC_{Doc}$ \\ [2.5pt]
      $ DGGen:\ DM_F\times V\times C_{Doc}\times RC_{Doc}\rightarrow DG\times C_{Doc}$ \\ [2.5pt]
      $ CGGen:DG\times KN\times C_{Doc}\times{RC}_{Doc}\rightarrow Doc_{KN}$ \\ [2.5pt]
      $ CFraming:{Doc}_{KN}\times KN\rightarrow{Doc}_{KN}$ \\ [2.5pt]
      $ SemGrounding:Doc\times KN\rightarrow Doc_{KN}$ \\ [2.5pt]
      $ Extr\left(f,dm_f\right)=Segmentation\circ Classification\circ Structuring$ \\ [2.5pt]
      $ Segmentation:F\times DM_F\rightarrow E$ \\ [2.5pt]
      $ Classification:E\times DM_F\rightarrow B\times L$ \\ [2.5pt]
      $ Structuring:B\times L\times DM_F\rightarrow H$ \\ [2.5pt]
      $ LingSimp\left(h\right)=SentenceSegmentation\circ NLPAnalysis\circ LinguisticAnalysis\circ SimplicityTransformation$ \\ [2.5pt]
      $ SentenceSegmentation:\ H\ \rightarrow\ Y$ \\ [2.5pt]
      $ NLPAnalysis:Y\rightarrow Y^\prime$ \\ [2.5pt]
      $ LinguisticAnalysis:Y^\prime\rightarrow Y^{\prime\prime}$ \\ [2.5pt]
      $ SimplicityTransformation:Y^{\prime\prime}\rightarrow U$ \\ [2.5pt]
      $ ConceptDetect\left(U,KN\right)=CandidateConceptDetection\circ CoreferenceResolution\circ RelationCuesDetection$ \\ [2.5pt]
      $ CandidateConceptDetection:U\rightarrow U^\prime$ \\ [2.5pt]
      $ CoreferenceResolution:U^\prime\times KN\rightarrow U^{\prime\prime}\times C_{Doc}$ \\ [2.5pt]
      $ RelationCuesDetection:U^{\prime\prime}\times C_{Doc}\rightarrow V\times C_{Doc}\times RC_{Doc}$ \\ [2.5pt]
      $ DGGen\left(DM_F,V,C_{Doc},RC_{Doc}\right)=DiscourseUnitsDetection\circ DiscourseRelationsDetection$ \\ [2.5pt]
      $ DiscourseUnitsDetection:DM_F\times V\times C_{Doc}\rightarrow DU$ \\ [2.5pt]
      $ DiscourseRelationsDetection:DM_F\times V\times{RC}_{Doc}\rightarrow DE$ \\ [2.5pt]
      $ DG=\left(DU,DE\right)$ \\ [2.5pt]
      $ CGGen\left(DG,KN,C_{Doc},RC_{Doc}\right)=SentenceGraphGeneration\circ SentenceGraphAggregation$ \\ [2.5pt]
      $ SentenceGraphGeneration:DG,KN,C_{Doc},RC_{Doc}\rightarrow DG\times ED_{Doc}$ \\ [2.5pt]
      $ SentenceGraphAggregation:DG\times ED_{Doc}\rightarrow Doc_{KN}$ \\ [2.5pt]
      $ Doc_{KN}=(F,DM_F,DG,C_{Doc},ED_{Doc})$ \\ [2.5pt]
      $ CFraming\left(Doc_{KN},KN\right)=FrameIdentification\circ KnowledgeScoping$ \\ [2.5pt]
      $ FrameIdentification:ED_{Doc}\times KN\rightarrow{Doc}_{KN},X_{Doc}$ \\ [2.5pt]
      $ KnowledgeScoping:{Doc}_{KN}\times KN\times X_{Doc}\rightarrow{ED}_{Doc}$ \\ [2.5pt]
      $ SemGrounding\left({Doc}_{KN},KN\right)=ConceptGrounding\circ RelationsGrounding\circ FramesGrounding$ \\ [2.5pt]
      $ ConceptGrounding:C_{Doc}\times KN\rightarrow C_{Doc}$ \\ [2.5pt]
      $ RelationsGrouding:{ED}_{Doc}\times KN\rightarrow{ED}_{Doc}$ \\ [2.5pt]
      $ FramesGrounding:{ED}_{Doc}\times KN\rightarrow{ED}_{Doc}$ \\ [2.5pt]
      \hline
    \end{tabular}
  \end{adjustbox}
  \vspace{2.5pt}
  \caption{Compendium of Gyan Platform Operations}
\end{table}

\subsubsection{\textbf{Discourse Model Identification}}

Gyan’s \textbf{$DMID$} or \textbf{Discourse Model Identification} takes a file $f \in F$ and the knowledge base of all the Discourse Models ($DM$) from the Knowledge Net ($KN$) and identifies a single discourse model $DM_f$ that the document in the file belongs to.

\begin{center}
  $ DiscourseModelId:F\times DM_{KN}\rightarrow DM_f $
\end{center}

The determination of the correct ${dm}_f$ involves running the file $f$ through all the identification rules ${id}_{i,j}\in{DModel}_{ID,{dm}_j},\ \forall\ {dm}_j\in{DM}_{KN}$, and finding the most appropriate ${dm}_f$.

\subsubsection{\textbf{Extraction}}

Gyan’s $Extr : F \times DM_f \rightarrow H$ or Extraction step is a function that maps a physical file $F$ into an abstract document representation $H$, by leveraging the parsing rules from the document’s discourse model ${DM}_f$.  $F$ can be of any file-format (pdf, docx, html, etc.), but $H$ is an extensive representation (gyan-txt format) that can represent information from any type of content.

A physical textual document $f \in F$ is a finite, ordered set of primitive graphical elements $e \in GE$.  Therefore, a physical file $f$ with $n$ graphical elements can be represented as $f = \{e_1, e_2, \ldots, e_n\}$.

Each graphical element $e \in E$ is defined by the tuple $e_i = \{co_i, pos_i, \widehat{et_i}, attr_i\}$, where $co_i$ represents the contents of the element, $pos_i$ represents the physical position of the element within $f$, $\widehat{et_i} \in \widehat{ET}$ represents the element type of this element, from a list of fixed element types (e.g. span, pdf block, SVG instructions, div tag, etc.) and $attr_i$ is collection of $k$ attributes $\{attr_{i_1}, attr_{i_2}, \ldots, attr_{i_k}\}$ where every $attr_{i_j}$ is a tuple $\{key_{i_j}, value_{i_j}\}$ where $key_{i_j} \in K_{attr}$ comes from a fixed set of attribute keys.

The abstract document representation $h \in H$ is a hierarchical tree-structure $h = (N, E)$.  $N$ is a set of nodes $n_i$, represents a logical document block (e.g. paragraph, table, section, etc.).  $E$ is a set of directed edges that define parent-child relationships between these logical document blocks, including their sequence and orders.

Every logical document block $n_i \in N$ is defined by the tuple $n_i = (b_i, \widehat{l_i})$, where $b_j$ is a sequence of the native graphical elements $e \in E$ from the physical file $f$ and $\widehat{l_i} \in \hat{L}$ is a label from set of all the various types of logical document blocks like $\_section$, $\_paragraph$, $\_table\_cell$, $\_bullet\_point$, etc.

The $Extraction(f) = (Segmentation \circ Classification \circ Structuring) (f)$ function is made up of three sub-functions.  $Segmentation : F \times DM_F \rightarrow E$ takes the physical document $f$ and extracts graphical elements $e_i \in E$, while leveraging the parsing rules from the discourse model $DM_F$ of the file.  $Classification : E \times DM_F \rightarrow B \times L$ leverages the rules from the discourse model of the document $DM_F$ and groups the extracted graphical elements into a sequence of logical document block $b_i \in B$ and provides it a label $\widehat{l_i} \in \hat{L}$, thereby creating the nodes of the abstract document representation $h$.  $Structuring : B \times L \times DM_F \rightarrow H$ algorithm them analyses these logical document blocks and utilizing the specifications within the discourse model $DM_F$ of the document, organizes them into a tree hierarchy.

\subsubsection{\textbf{Linguistic Simplification}}

Gyan’s $LingSimp: H \rightarrow U$ or the \textbf{Linguistic Simplification} is a function that takes in the abstract document representation $h \in H$ and creates a list of simplified sentence representations $U$ by applying iterative, rules-based linguistic transformations.  The function comprises of 4 sub-steps, defined by:

\begin{center}
	$LingSimp(h) = SentenceSegmentation\ \circ\ NLPAnalysis\ \circ\ LinguisticAnalysis\ \circ\ SimplicityTransformation$
\end{center}

The $SentenceSegmention : H \rightarrow Y$ function traverses the abstract document representation $h \in H$ in the reading order and creates the text content ${tco}_i$ for each block $n_i \in N$ by concatenating the ${co}_i$ from every graphical element $e \in E$ from its logical document block $b_j$.  The output is an ordered list of sentences $Y=\{y_1, y_2, \ldots, y_p\}$.  Each sentence $y_k \in Y : y_k = (t_k, src_k)$ where $t_k$ is the raw text content of the sentence and $src_k$ contains the reference to the $blocks b_j$, the graphical elements $e \in E$ resulted in the construction of this block.  This ensures the complete traceability of every sentence to the physical location in the file.

The $NLPAnalysis : Y \rightarrow Y^\prime$ function applies a series of standard NLP steps to every  $y_k \in Y$ to produce an enriched sentence ${y_k}^\prime \in Y^\prime$.  Each ${y_k}^\prime = (y_k, A_k)$ contains the original sentence $y_k$ and $A_k = (Tokens, POSTags, NamedEntities , DependencyParseTree, \ldots)$, which are a set of NLP-annotations.  Each annotation is produced by employing a standard, licenced, open-source and state-of-the-art annotator available in the open-source community, with pre-processing simplification, pre-substitution, post-substitution and post-processing correction rules, which have been identified to fix some of the common errors committed by these annotators, by reviewing their source code, training data distributions and performance on hand-crafted datasets.

The $LinguisticAnalysis : Y^\prime \rightarrow Y^{\prime\prime}$ function identifies grammatical categories and other advanced linguistic features by using the basic NLP annotations and curated Linguistic Rules for each feature.  Every ${y_k}^{\prime\prime} \in Y^{\prime\prime}$ is defined by the tuple ${y_k}^{\prime\prime} = {(y_k}^\prime,L_k)$ where ${y_k}^\prime \in Y^\prime$ is the enriched version of the sentence $y_k \in Y$ and $L_k = (Voice, Animacy, Case, Definiteness, Gender, Modality, Aspect, \ldots)$ is a set of linguistic attributes within the sentence ${y_k}^\prime \in Y$.

The $SimplicityTransformations : Y^{\prime\prime} \rightarrow U$ function applies a set of ordered transformation rules $R = \{r_1, r_2, \ldots\}$ to each sentence ${y_k}^{\prime\prime} \in Y^{\prime\prime}$.  This step is iterative; implying that every transformation application by a rule $r_i \in R$, triggers a reapplication of transformation rules right from $r_1 \in R$.  The output is a final set of simplified sentences $U = \{u_1, u_2, \ldots, y_\Phi\}$ where the total number of sentences $\Phi$ may be greater than $p$.  The rules handle important constructs in language like idioms, including others.

\subsubsection{\textbf{Concept Detection \& Relation Cues Identification}}

Gyan’s $ConceptDetection:U\times KN\rightarrow(V,C_{Doc},{RC}_{Doc})$ function takes the simplified sentences $u_i\in U$ and creates $c_i\in C_{Doc}$ within them, groups them into sets of same-concept referring concepts  $n_{c_i}\in N$ and collects the remaining sentence segments ${rc}_i\in RC$ as Relation Cues.  The $ConceptDetect = CandidateCDetection \circ CorefRes \circ RelationCuesDetect$ works in three sub-steps.

The $CandidateCDetection:U\times KN\rightarrow U^\prime\times C_{Doc}$ function takes the simplified sentences $U=\{u_1, u_2, \ldots, u_\Phi\}$ and the normalized concepts $N \in KN$ and creates a set of annotated sentences $U^\prime = \{u_1^\prime, u_2^\prime, \ldots, u_\Phi^\prime\}$, where every $u_k^\prime\in U^\prime$ is defined by the tuple $u_k^\prime=(u_k, C_k)$ containing the original simplified sentence $u_k$ and a set of all the candidate concepts  $C_k$.  The candidate concepts space of the document $C_{Doc} = C_1 \cup C_2 \cup \ldots \cup C_\Phi$ is the union of all the concepts $C_i$ from all the $\Phi$ simplified sentences from the document.  The concepts $c_i$ are not just entities, but also include other nouns, noun-phrases and other compound predicates, based on rules on simple NLP annotations and advanced Linguistic annotations.  Every $c_i \in C_{Doc}$ is a tuple of $c_i = \{c_{name}, c_{norm}, c_{attr}, c_{trace}\}$, as defined in the Abstract Document Model and in this step, the $c_{norm}$ values are not calculated.

The $CorefRes : U^ \prime \times C_{Doc} \times KN \rightarrow C_{Doc}$ function takes $U^\prime=\{u_1^\prime, u_2^\prime, \ldots, u_\Phi^\prime\}$, the normalized concepts $N \in KN$ and the previously extracted $C_{Doc}$ concepts and updates the $c_{norm}$ value of every concept $c_i \in C_{Doc}$, either with a value $n_i \in N$ or to $null$.  Note that multiple concepts $c_i \in C_{Doc}$ could refer to the same normalized concept in $N$.  The $CorefRes$ function employs a series of rules on basic NLP and advanced linguistic annotations to determine which concepts are referring to the same concept.  These rules may leverage the concept specification within $N \in KN$ to infer more complex ways in which multiple concepts in the document infer to the same concept in $N$.

The $RelationCuesDetect : U^\prime\times C_{Doc} \times KN \rightarrow V \times C_{Doc} \times RC_{Doc}$ takes all the concept-detected sentences $U^\prime = \{u_1^\prime, u_2^\prime, \ldots, u_\Phi^\prime\}$, the document concepts $C_{Doc}$ and leverages the Knowledge Net to generate sentences $V=\{v_1, v_2, \ldots, v_\Phi\}$.  Every sentence $v_i \in V$ is made up of a tuple $v_i = (u_i^\prime,rc_i)$, where $u_i^\prime$ is the concept-detected sentence and $rc_i\ \in RC$ is a set of leftover phrases and sentence segments.  The set $RC = {rc}_1 \cup {rc}_2 \cup \ldots{rc}_\mathrm{\Phi}$ is the whole document’s set of relation cues.

\subsubsection{\textbf{Discourse Graph Generation}}

Gyan’s $DGGen:(DM_F, V, C_{Doc}, RC) \rightarrow DG$ or the \textbf{Discourse Graph Generation function} takes the analyses from the previous steps and applies the rules from the document’s discourse mode $DM_F$ to create a discourse graph $DG = (DU, DE)$, where $DU$ is the set of all the discourse units within the document and $DE$ is a set of all the discourse edges between a pair of discourse units $du_i, du_j \in DU$.  The $DGGen= DUDetection \circ DRDetection$ function contains two sub-steps.

The $DUDetection:DM_F\times V \times C_{Doc}\rightarrow DU$ function performs the detection of various discourse units within the document, by applying discourse units detection rules from the document’s discourse model $DM_F$.  $DU=\{du_1,du_2,\ldots,du_z\}$ is the set of all the identified discourse units within the document.

Each $du_i\in DU$ is a tuple $du_i=(name_{du_i}, attr_{du_i}, V_{du_i}, C_{du_i})$
\begin{quote}
  where:
  \begin{itemize}
    \item  $name_{du_i}$ is the name of the discourse unit $du_i$ as described in the $DM_F$.  Examples include $Idea$, $Enumeration$, $Section$, etc.
    \item  $attr_{du_i}=\{kv_1, kv_2, \ldots, kv_n\}$ is a set of key value pairs where keys are specific attributes related to the discourse unit, ad defined in the discourse model of the document $DM_F$.
    \item  $V_{du_i}=\{v_1, v_2, \ldots, v_\alpha\}$ is a subset of $V$ which are the simplified, concepts- and relation-cues-determined sentences from the document.
    \item  $V=\{V_{du_1}\cup V_{du_2}\cup\ldots\cup V_{du_\alpha}\}$, implying that every sentence must be a part of at least one $DU$.
    \item  $\forall du_i \in DU : \left|v_{du_u}\right| \neq 0$, implying that no discourse unit can be made up of no sentences.
    \item  $\forall du_i \in DU \forall v_j, v_k \in V_{du_i} \colon \left| v_j \right| \not\equiv \left|v_k\right| $ implying that there is no assumption on the equality of number of sentences within any two discourse units of a document.
    \item  $C_{du_i}=\{c_{du_i1},c_{du_i2},\ldots,c_{du_i\beta}\}$ is a subset of all the concepts form the document $C_{Doc}$ that belong to the sentences $V_{du_i}$.
  \end{itemize}
\end{quote}

In a document belonging to a typical textual discourse model, DUDetection starts by assigning every simplified sentence $v_i\in V$ as a discourse unit $du_i\in DU$ of $name=Idea$.  Then, performs multiple iterations where it groups multiple discourse units from an iteration into a composite discourse unit, based on either mechanics-oriented rules or linguistics-oriented rules.  These discourse units are then passed-on as inputs to the next iteration, until the output discourse units from every iteration are not the same as the input discourse units from that iteration.

The $DRDetection:V\ \times C_{Doc}\ \times DU\rightarrow DE$ function takes the discourse units created from the DUDetection step and applies rules in $DM_F$ to determine a set of edges $DE$.

Every $de_i\in DE$ is a tuple $de_{i,j}=(du_i,du_j,name_{i,j},r_{DM_F},attr_{i,j})$

\begin{quote}
  where:
  \begin{itemize}
    \item  $du_i,du_j\in DU$ denotes the source and the target of the edge $de_{i,\ j}$.
    \item  $name_{i,j}$ is the name of the edge, as specified in $DM_F$.
    \item  $r_{DM_F}$ is the relationship between the two discourse units, as specified in $DM_F$.
    \item  $attr_{i,j}$ is a set of key-value pairs where the space of keys is as defined in the $DM_F$ and stores the values of these attributes for this instance of the discourse edge.
  \end{itemize}
\end{quote}

In a document belonging to a typical textual discourse model, the attributes $attr_{i,j}$ stores information related to structure and reading order with values like $Hierarchy\_Parent$, $Hierarchy\_Child$, $Preceding$, $Structural\_Coreference$, etc.  The relation $r_{DM_F}$ stores the rhetorical relationship like $Elaboration$, $Justification$, $Background$, $Conclusion$, etc.

\subsubsection{\textbf{Concept Graph Generation}}

Gyan’s $CGGen:DG\times KN\times C_{Doc}\times RC_{Doc}\rightarrow{Doc}_{KN}$ or the Concept Graph Generation function takes the discourse graph $DG=(DU, DE)$ and converts it into a Document SKG $Doc_{KN} = (F, DM_F, DG, C_{Doc}, ED_{Doc})$.  The function works by creating a concept graph of every sentence $s_i\in V_j$ where $V_j$ is the set of sentences of the discourse unit $du_j$, and then aggregating it to form $CG_j = (CON_j, ED_j)$ or a Concept Graph for every discourse unit $du_j \in DU$.  $CGGen = SCGGen \circ SCGAggr$.

The $SCGGen : DU \times KN \times C_{Doc} \times RC_{Doc} \rightarrow DU \times ED_{Doc}$ or the Sentence Concept Graph Generation function works on every sentence $v_{ij} \in V_i$ of all the sentences of the discourse unit $du_i$ for all $du_i \in DU$ and generations a ${CG}_{ij} = (CON_{ij}, ED_{ij})$ where the $c_k \in CON_{ij}$ are a set of all the concepts in the sentence $v_{ij}$ and the $ed_{k,m} \in ED_{ij}$ is an edge between the concept $c_k$ and $c_m$.

Every $c_k \in CON_{ij} \subset C_{Doc}$ is a tuple $\{c_{name}, c_{norm}, c_{attr}, c_{trace}\}$, as described in the Abstract Document Graph.  Every edge $ed_{k,m} \in ED_{ij}$ is represented with the quadruple $ed_{k,m} = (c_k, c_m, r_{norm}, x, x_{norm}, type)$, as described in the Abstract Document Graph.  In this step, the $x$ and $x_{norm}$ are left unassigned and are populated in the Contextual Framing step.  The $r_{norm}$ and type of every $ed_{k,m}$ is evaluated from the relation cues $rc_{ij} \in RC_{Doc}$ and is mapped to one of the relations $R$ from the Knowledge Net $KN$.

The $SCGAggr : DG \times ED_{Doc} \rightarrow Doc_{KN}$ or the Sentence Concept Graph Aggregation function works on every discourse unit $du_i \in DU$ of the document and aggregates the concept graph $CG_{ij}$ of every sentence $v_j \in V_i$ in $du_i$ and aggregates it into a consolidated concept graph $CG_i = (CON_i, ED_i)$ of the discourse unit $du_i$.  The process involves taking a semantic union of all the nodes and edges of the individual sentence concept graph.  $CON_i = CON_{i1} \cup_{sem} CON_{i2} \cup_{sem} \ldots \cup_{sem} CON_{in}$ where $CON_{ij}$ are the concepts from the concept graph of sentence $v_{ij} \in V_i$ of the discourse unit $du_i$.  Similarly, $ED_i = ED_{i1} \cup_{sem} ED_{i2} \cup_{sem} \ldots \cup_{sem} ED_{in}$.

\subsubsection{\textbf{Contextual Framing}}

At this stage, the Document Semantic Graph $Doc_{KN}$ is a Discourse Graph DG as the first layer and every discourse unit $du_i \in DU$ in the $DG$ has a concept graph $CG_i$.  Each concept graph has concepts normalized to a $n_k \in N \in KN$ and each edge has relations normalized to $r_j \in R \in KN$.  The missing annotations include the $x$ and $x_{norm}$ which are the contextual frames of every relationship within the discourse graph and concept graph of every discourse unit of the discourse graph.

Gyan’s $CFraming : Doc_{KN} \times KN \rightarrow Doc_{KN}$ of the Contextual Framing function takes assigns the contextual frame $x$ to every discourse and concept edge in $Doc_{KN}$ and maps them to the contextual frames space $X$ in the $KN$ using linguistic rules and algorithms.  The function works by identifying all the unique contextual frames $x_i$ within the document $X_{Doc}$ by with respect to the contextual frames within the Knowledge Net and then assigns a frame to every discourse and concept edge.  These two sub-steps are modelled as:  $CFraming(Doc_{KN},KN) = FrameIdentification \circ KnowledgeScoping$.

The $FrameIdentification : Doc_{KN} \times KN \rightarrow Doc_{KN}$ or the Frame Identification function works on all edges $de_i \in DE$ and $ed_{mn} \in ED_n \in du_n \in DU$, inspects all the linguistic features and populates a $X_{Doc}$ with $\{x_1, x_2, \ldots, x_\epsilon\}$ where every $x_i$ is a tuple of $\{x_{i,norm}, x_{i,defs}\}$.  The $x_{i,defs} = \{kv_1, kv_2, \ldots, kv_\gamma\}$ is a collection of key-value pairs where the keys belong to a fixed set of primitive frames (e.g. $time$, $location$, $aspect$, $modality$, etc.) and the values are explicit permissible values for each of those primitive frames.  The $x_{i, norm}$ is a mapping of $x_i$ to one of the frames $x_j \in X \in KN$.

The $KnowledgeScoping : Doc_{KN} \times KN \times X_{Doc} \rightarrow Doc_{KN}$ or the Knowledge Scoping function takes all the identified and mapped frames $X_{Doc}$ and populates the $x$ and $x_{norm}$ values in every $de_i \in DE$ and $ed_{mn} \in ED_n \in du_n \in DU$.

\subsection{Semantic Grounding of a Document GMR in Gyan}

The Gyan $Encoder$ converts a file into a Document GMR $Doc_{KN}$ in the context of the provided knowledge net $KN$.  This step continuously grounds the document to the $KN$ with each and every step.  This is known as \textbf{continuous semantic grounding} of the Document.  However, depending on a use-case, it may be beneficial to encode the file into a document SKG using a null knowledge net $KN_\Phi$ and then apply a knowledge net post-facto, a process known as \textbf{post-hoc semantic grounding} of the document which is achieved through additional grounding functions in Gyan.

Gyan’s $SemGrounding:Doc_{KN_\Phi} \times KN\rightarrow Doc_{KN}$ or the \textit{semantic grounding function} takes a document SKG $Doc_{KN_\Phi}$ that was grounded on a null knowledge net $KN_\Phi$ and maps its concepts, relations and frames from the $Doc_{KN_\Phi}$ to those of the provided knowledge net $KN$.  $SemGrounding = ConceptGrounding \circ RelationsGrounding \circ FramesGrounding$.

Each of the three grounding steps map the concepts, relations and frames from the document space to the space of the Knowledge Net $KN$.  These steps utilize the ideas from the coreference resolution, concept linking, concept normalization and concept equivalence to map the components of the processed documents from one knowledge net (null knowledge net, in this case) to another.

While a post-hoc grounding is fast, it is pertinent to note that performing a post-facto grounding of a document’s GMR comes with some limitations as there’s a collection of rules that can only be applied at the time of creating fundamental structures of the document’s GMR.  In order to get past this limitation, the whole file can be subjected to continuous grounding through the $Encoder$ with the provided $KN$.  The trade-off is between time and completeness.

\subsection{Knowledge Discovery \& Building the Knowledge Net}

The $KnowledgeDiscoverer$ (or $KD$ or just $Discoverer$) performs the knowledge discovery and creates one or more Knowledge Net, including the Universal Knowledge Net.

The fundamental Knowledge Discovery steps performed by the $Discoverer$ are:
\begin{enumerate}
	\item  Starts with an input concept and a corpus
	\item  processes the documents in the corpus using the Meaning Encoder,
	\item  finds all the occurrences of the input concept,
	\item  classify each of the occurrences of these concepts into whether they are the dominant topics from the corresponding Discourse Units; a process we term as relevance determination
	\item  extracts the relation along with the context in which the relations always apply.
	\item  determination of whether the context is generic enough to occur in other cases
	\item  saves the relations in the knowledge net.
\end{enumerate}

This method results in a Knowledge Net that is grounded on the provided corpus.  The corpus can be licensed content or freely available free-for-commercial-use content or content accumulated by an Enterprise over several years of operations.  This way, an Enterprise Knowledge Net could be constructed in a manner such that its nuances can be captured in it.  This also provides a way to Enterprises to attain an edge over other enterprises.

This fundamental process of knowledge discovery enables the following methods to comprehensively build complex and comprehensive knowledge nets:
\begin{enumerate}
	\item  Build a knowledge net for a vocabulary of terms
	\item  Discover vocabulary from a corpus of documents and iteratively build a knowledge net.
	\item  Build a knowledge net from a series of provided dictionaries.
	\item  Incrementally, add discover and add a knowledge for a new concept to an existing knowledge net.
	\item  Add knowledge from a corpus of documents to an existing knowledge net
	\item  Add a knowledge net to an existing knowledge net
\end{enumerate}

Some of the common utilities within the $Discoverer$, following components can make the task of creating knowledge nets easier:
\begin{enumerate}
	\item  $DiscoverVocabulary$ utility to identify important concepts from a corpus of documents.
	\item  $AggregateCorpus$ utility helps aggregate a corpus for a given concept from the web, by firing searches on a web search engine and collecting the documents from the sources that fulfill a certain source credibility criteria.
	\item  $DetermineRelevance$ helps to identify documents and discourse units within the documents that presents a given concept as an important topic of discussion in them.
\end{enumerate}

\subsubsection{Gyan Base Knowledge Net}

Gyan’s Base Knowledge Net is a repository of knowledge that tries to contain the full breadth of concepts in the world, with some/little depth.  The process used to create the base knowledge net are:
\begin{enumerate}
  \item Vocabulary was seeded from WordNet, KBPedia, ConceptNet, Cambridge English Dictionary and Oxford English Dictionary.
  \item Concepts in the Vocabulary from various sources were disambiguated and linked to ensure that the senses are maintained.
  \item Relations from KBPedia and ConceptNet were imported into the Base Knowledge Net.
  \item Definitions were discovered for the concepts and were processed using the Meaning Encoder to add additional fundamental knowledge (e.g. hypernymy).
  \item Wikipedia and Encyclopedia were sourced and converted into a corpus of HTML documents.  A generic HTML Discourse Model was used to parse and pre-process all the contents into the documents.
  \item Meaning Encoder was run on all the documents in the corpus and for every term in the vocabulary, relations were extracted from the relevant discourse units and saved in the Base Knowledge Net.
\end{enumerate}

\section{Knowledge Stores}

Section 1.3.2 and 1.3.3 defines the fundamental difference between a Knowledge Net and Knowledge Store.  Knowledge Stores are repository of Meaning Graphs and Knowledge from a collection of underlying documents, that can be queries on.  Although a Universal Knowledge Net, a massive knowledge net introduced in 1.2.2, is the ideal way to simulate the knowledge and intelligence of the whole world, smaller and more practical knowledge nets have been proven to be completely useful to solve niche and domain or topic restricted problem.  Our base knowledge net becomes a good and effective starting point for creating a knowledge net or knowledge store that covers and contains all the background knowledge for a given collection of documents.

\subsection{Gyan Foundation Knowledge Stores}

Gyan provides some foundation knowledge stores out of the box, that can either be used for a domain or can be used as a starting point to create a knowledge net or knowledge store for a deeper subject area.

\subsubsection{Life Sciences Knowledge Store}

Gyan’s Life Sciences Knowledge Store \textit{LS-KS} is a store that is being used by the Life Sciences community to interact with the knowledge of most of the sub-fields of Life Sciences.  The Life Sciences Knowledge Store was created using the following approach:

\begin{enumerate}
  \item  Seeded the LS-KS with the Base Knowledge Net.
  \item  Curated a vocabulary of important terms from the various sub-fields of life sciences.
  \item  Curated a list of encyclopedias and credible reference information for the various fields of Life Sciences.  E.g. PubChem, Cleveland Clinic, etc.
  \item  Curated a list of all the research papers (full or abstract) for the various fields of Life Sciences.  E.g. PubMed, Science Direct, etc.
  \item  Process the content and iteratively create a knowledge store.
\end{enumerate}

The \textit{LS-KS} is available as a Sandbox and the access can be obtained by contacting the authors.  The Sandbox also illustrates the efficacy of the Gyan’s approach to AI by presenting question samples from the PubMedQA dataset – both labelled and unlabeled.

\subsubsection{USPTO Knowledge Store}

Gyan’s USPTO Knowledge Store \textit{USPTO-KS} is a store of all the US Patents and is being tested for use by Global Patent Research Analysts.  Since the source here is unfragmented, unlike in the \textit{LS-KS}, the USPTO-KS is built on top of the Base Knowledge Net with curated dictionaries and taxonomies of the various classification systems prescribed by the regulatory body.

\subsection{Enterprise Knowledge Store}

Enterprise Knowledge Store is a Knowledge Store that an enterprise builds by starting from the Gyan Base Knowledge Net and adding knowledge and information about concepts and data present in their ever-increasing enterprise repository.  Once created, these knowledge stores can be kept current by adding all incremental documents and information into it.  Given that the model can handle multiple modes of data (even within texts), all strategic, tactical, as well as operational data can also make it to the Enterprise Knowledge Store.  Several Enterprise-specific procedures can also be added to the Enterprise Knowledge Store to reflect a customized context.

\section{Inferencing in the Gyan Model}

\subsection{Query Model}

A query triggers \textit{Reasoning} in Gyan.  A query on a Knowledge Store is broken down into background, instruction, constraint concept graph and context concept graph.  A complex query with multiple sub-queries is worked on independently and then the responses are merged back based on the instructions.

\begin{itemize}
	\item
      \textbf{Background} serves as a transient increment to the documents in a Knowledge Store.
	\item
      \textbf{Instruction} can be classified into Retrieval, Analytical and Modification instructions.
	\item
      \textbf{Constraint Concept Graph} is the concept graph of the main question and Context Concept Graph is the concept graph of the context.  Constraints and Context are separated to make matching independent of the contextual connection as described in the query.
	\item
      \textbf{Retrieval Instructions} include Verify, Find, Count and Describe.
	\item
      \textbf{Analytical Instructions} include Compare, Aggregate, Find-Path and Summarize.
	\item
      \textbf{Modification Instructions} include instructions to Update the Knowledge-Store with a Transient Background of the Instruction.
\end{itemize}

A \textbf{complex task is a procedure} with a series of sequential sub-tasks, each of which is mapped to a particular instruction from Gyan’s Instruction Set.

Knowledge Store \textbf{augments the Instruction Set} by being able to query more procedures from either the \textit{Procedural Knowledge Net} of the Knowledge Store, or one derived from the \textit{Background} of the Query.  This is similar to the \textbf{extension interface} for working with other AI agents through standardized or non-standardized protocols like MCP.  Examples can be procedures to add two numbers or output a chunk of code in Java to sort a generic array.

The \textbf{outcomes of each of these Instructions} are at least the \textit{Knowledge Sub-graphs} and \textit{Discourse Unit Sub-Graphs} from the underlying \textit{Knowledge Stores}.  The \textbf{synthesis operation} involves Goal/Instruction specific presentation of the matched and returned sub-graphs.  The synthesis methods are also procedures that can be defined within the Knowledge Store to produce the outcome with various features (e.g. completely extractive).

As a specific example, the \textbf{MCQ Answering Procedure} involves creating one $FindPath$ Instruction for each option and then evaluating them based on the properties of the returned paths (if one exists) to find the best one.

\subsection{Graph Equality and Similarity}

Two graphs are isomorphic if they have the exact same nodes and edges through some mapping of nodes and edges of one graph to those of the other graph.  Two graphs are homomorphic the connected nodes in one are either connected nodes in the other or are the same node in the other.  A relational calculus applied on a graph produces a deductive closure of the graph.

If two graphs have an isomorphic deductive closure, they are logically equivalent.  In other words, two Graphs are logically equivalent if one can be transformed into another, through a series of transformations consistent with a Relational Calculus.

Computational Complexities of Isomorphism Detection is $O(m log n)$ for two graphs with $m$ nodes and $n$ edges, through the Weisfeiler-Leman (WL) Heuristic.  SOTA Complexity of Subgraph Isomorphism between two graphs of dimensions $(n, m)$ and $(N, M)$ is $O(N^n \times n^2)$.  Best known implementations of Graph homomorphism detection have a complexity of the order of $O(N^n)$.  Common subgraph isomorphism detection can only be completed in time proportional to $O(c^{n \times N})$, where $c$ is the average degree of the nodes in the two graphs.

Gyan implements computation of equality and various measures of similarity between two graphs very efficiently.  One of the two graphs is a query graph, and the other one is large list of graphs of discourse units.  The primary computation is to find all the discourse units that contain an isomorphic deductive closure of the query graph.  We define this kind of graph search as an \textbf{Inferential Sub-graph Search}.

The efficiency and scalability of Gyan’s implementation is a combination of powerful indexing mechanisms, the right amount of inference materialization, a fast implementation of a multi-pass refine and match steps and intelligent query-time inferencing.  Linguistic breakdown of the document into context-containing discourse units ensures that there are no run-away inference closure computation.  Logical segmentation break-down of a bigger graph into smaller sub-graphs (e.g. constraints and contexts) and graph kernels ensures fast retrieval of discourse units.

\subsection{Inferences and Reasoning}

Inferences are steps in reasoning.  Reasoning can be classified into four foundational methods: deductive (e.g. expert systems), inductive (e.g. machine learning), abductive (e.g. Bayesian networks) and analogical (e.g. case-based reasoning).  Deductive reasoning produces new relations.  Inductive reasoning produces new rules.  Abductive reasoning produces explanations.  Analogical reasoning produces new analogies.

Gyan implements the various types of reasoning through graph matching and graph projections throughout the various steps of the Platform.  Deductive reasoning involves finding graphs from the rules or queries.  Inductive reasoning involves finding similar graphs (observations) to create a generic knowledge graph or operate on them.  Abductive reasoning requires searching sub-graphs and then finding connections or paths or performing deductive and inductive reasoning on/between the returned results.  Analogical reasoning requires searching sub-graphs of the subjects (and their attributes) of comparison and then performing deductive and/or inductive reasoning on them.

\section{Gyan Platform Reasoner}

\subsection{Knowledge Store Enrichment}

Inference materialization is used to achieve heuristics-based completion of deductive closure.  More edges are added.  More rules are discovered, triggering addition of more edges.  Concepts are classified, clustered, etc.  Pre-computation of some properties like attributes, expansions, etc.

\subsubsection{Gyan Query Encoder}

Gyan’s $QueryEncoder:U \times KS \rightarrow Q (INS, BG, G_{Constraints}, G_{Context})$ query parser takes a user input $U$ and a Knowledge store reference $KS$ and produces a parsed query $Q$.  As described earlier, a parsed query contains a background $BG$, an Instruction $INS$, a concept graph of main constraints $G_{Constraints}$ and a concept graph of the contextual constraints $G_{Context}$ on the user input.  The user input is always converted into a text for the purpose of encoding.  Once converted into text, a pipeline similar to Gyan’s textual document encoder Encoder is run to encode the user input into $Q$.

\begin{center}
  $ QueryEncoder: U\times KS \rightarrow Q $ \\
  $ Q = (INS, BG, G_{Constraints}, G_{Context}) $ \\
  $ QueryEncoder(U, KS) = LingSimplify \circ InputSegment \circ InstructDetect \circ CGGen $
\end{center}

The main steps in $QueryEncoder$ includes $LinguisticSimplification$ on the contents of the input, followed by $InputSegmentation$, $InstructionDetection$ and $ConceptGraph$ generation.

\subsubsection{On-demand Knowledge Aggregation}

A Knowledge Store may not contain the knowledge or tools to follow a user’s instructions.  On-demand knowledge aggregation layers multiple knowledge stores in order to get access to knowledge and/or tools to execute a particular instruction.  It starts with the user’s knowledge stores, followed by the Enterprise knowledge store, one of the various Gyan’s foundational knowledge stores and ultimately to a \textbf{dynamic web knowledge store}.

The generation of dynamic web knowledge store starts with intelligently breaking the user’s query into search engine queries and identifying a set of focused high-authority sources that may contains information or tools to achieve the goal.  The process also allows the users to define their own list of sources for a particular instruction.

The subsequent step then fires the queries on the search engines on the specified or identified sources and collects all the search results.  The selection of articles in this phase focuses on picking top $N$ articles, but ensures that $N$ is not too large, given the well-known problem of rapidly dropping relevance of articles from search engine results after a particular ranking.

To ensure a good initial crop of relevant articles are accumulated, the intelligent search engine query formation incorporates a bunch of expansions so that even with a small $N$, enough articles are accumulated and cover all aspects of the user’s instructions.  Furthermore, a relevance algorithm assigns a degree of relevance to every article returned against the search query.  Selecting the articles above a threshold of degree of relevance also ensures that the articles are on-point.  A de-duplication step removes articles that contain redundant information.

Once a good batch of articles is collected, a dynamic knowledge store containing the knowledge and information from these documents is created.  The Knowledge Net required to accurately mine the knowledge from these documents in the user’s context is derived from one or more Knowledge Stores that the user has access to.

This dynamic knowledge store is then used in the knowledge store chaining to identify the knowledge, procedures and tools to perform the specified task.  Along with the query for which the dynamic knowledge store was generated, the user can fire other queries on this newly created dynamic knowledge store as well.

These dynamic knowledge stores created by the users for one or more of their queries can be curated by the user to grow the context the user wants to create for their queries.  Important or relevant documents from these dynamic knowledge stores can be added back to other knowledge stores by the user.

\subsection{Synthesizer}

As we will see in the following sections, all the reasoning tasks involve solving instances of \textit{Inferential Sub-graph Search}.  After the reasoning engine performs its graph matching operations, the raw result is a collection of subgraphs or Discourse Units. The Synthesizer is the final component in the query pipeline responsible for transforming this structured data into a final answer, formatted according to the initial Instruction.

We define $Synthesizer$ as a set of instruction-specific functions. Therefore:

\begin{center}
  $Synthesize_{Instruction}:Set<InferentialGraphIsomorphism> \times Verbosity \rightarrow Answer$.
\end{center}

The $Answer$ is a human-readable output that either performs the instructions or performs them and acknowledges or presents the outcome.

The $Synthesizer$ also takes a $Verbosity$ parameter that dictates the length or the number of details in the output.  For example, in a $Verify$ instruction, the shortest answer could be a simple and succinct $True$, but a more verbose answer could have details about the concepts and their relationships being subjected to verification.

\subsection{Retrieval Tasks}

\subsubsection{Verify}

The $Verify$ instruction is a retrieval task used to confirm whether a specific statement is true, based on the knowledge store. It is the formal equivalent of a \textit{yes/no} question.

A Verify query is defined by:

\begin{center}
  $Q_{KN} = (INS, G_{Constraint}, G_{Context})$
\end{center}

\begin{quote}
  where:
  \begin{itemize}
    \item  $INS$ = $Verify$,
    \item  $G_{Constraints}$ is the fully instantiated concept graph of facts to be verified
    \item  $G_{Context}$ defines the context and scope of the verification.
  \end{itemize}
\end{quote}

Examples:  \textit{Does Mitochondria play a role in remodeling lace plant leaves during programmed cell death?}

Gyan Reasoner takes the $G_{Constraints}$ and $G_{Context}$ and performs a inferential subgraph isomorphism on the knowledge store.  The search is successful if it finds at least one discourse unit whose deductive closure has a isomorphic graph of the constraints and context from the query.

$Synthesizer_{Verify}$ for this task could output a True or a False, or a verbose paragraph like:  \textit{"I found 12 instances in the knowledge store that confirm that Mitochondria play a role in remodeling lace plant leaves during programmed cell death."}  On another spectrum, it will explain the $G_{Constraint}$ by adding details on why, how etc.

\subsubsection{Multiple Choice Questions}

The $SelectBest$ instruction is a bit more complex than the plain Verify instruction because it requires additional processing after a bunch of Verify instructions of selecting the best option for the question.  The $INS=(SelectBest,[Option-A, Option-B, \ldots])$ and contains the references to all the options for the MCQ question.  $The G_{Constraints}$ contains the stem of the question and the $G_{Context}$ is derived from the question itself.

Gyan Reasoner breaks a SelectBest questions into as many Verify sub-queries as there are options in the MCQ.  This is followed by performing the individual verification tasks.  All the verification tasks, along with the graph matches, also return a plausibility score.  This plausibility score is a combination of several metrics and measures of the underlying inferential subgraph isomorphism.  The final answer is selected as the option with the best plausibility score.

$Synthesizer_{SelectBest}$ could simply output the final option or provide additional details around the selected option or every go over and beyond to explain why the other options were incorrect, depending on the Verbosity setting.

\subsubsection{Find}

The $Find$ instruction a core retrieval task that is used to find the value of an unknown variable or $INS=(Find,[Var1,Var2,\ldots])$.  The $G_{Constraint}$ is a partially instantiated concept graph and the $G_{Context}$ is a fully instantiated concept graph to define the scope of the query.

Gyan Reasoner again solves this through the inferential subgraph isomorphism with the partially instantiated $G_{Constraint}$ and the fully instantiated $G_{Context}$ concept graphs and returns all the matching sub-graphs and the discourse units.  The Reasoner then performs projections on the returned sub-graphs and identifies the values of the variables.  It also provides computes a plausibility score of the correctness of the output, based on various factors, including the trustworthiness of the source document(s) and the number of instances when the values of the variables that complete the sub-graph.

The $Synthesizer_{Find}$ could either output the values of the variables or the verbatim discourse-unit(s) or an assertive statement with the answer or details of the concepts in the question or any combination of the above.

\subsection{Analytical Tasks}

\subsubsection{Compare}

The $Compare$ instruction is a high-level analytical task.  Gyan Reasoner doesn't find a single pattern but instead performs a differential analysis between two or more entities based on the information in the knowledge store:

\begin{center}
  $ INS=(Compare,[ConceptA,ConceptB,\ldots])$
\end{center}

The $G_{Constraints}$ contains the fully instantiated concept graphs of the concepts to be differentiated and $G_{Context}$ might contain the scope of the concepts or comparison or both.

Gyan Reasoner takes the concept graphs (constraints and contexts) of the concepts to be compared and creates a concept SKG of all the concepts that satisfy the constraints and are scoped by the context.  The concept SKG contains all the details of the concepts within the Knowledge Store.  The Reasoner, then, performs a differential analysis (another form of inferential subgraph isomorphism) on similar aspects of the concepts to determine whether they are similar or different.  The information here is a series of similarities and differences between the concepts.

The $Synthesizer_{Compare}$ in the simplest form could present a table with rows as the attributes and columns as the concepts.  The values in every sell of this table could just be other concepts, full discourse units or elaborate details about the aspect and their relationship to the concepts.  On the other hand, the output could be paragraphs of text describing these similarities and differences either as running text or hierarchical enumerations with or without sections and sub-section.

\subsubsection{Find-Path}

The $FindPath$ instruction is an analytical task used to discover the relationship between two concepts that may not be directly connected.  For the FindPath instruction, the $INS=(FindPath,[ConceptA,ConceptB])$.  The $G_{Constraints}$ typically contain the concept graphs of the two concepts and the $G_{Context}$ contain all the scopes and constraints either on the concepts or the required paths between them.

Depending on the instruction, the instruction requires the Reasoner to find the most salient or shortest or longest chain of connections between them on one end and all the paths between them at the other end.  It may also require finding paths that contain certain other concepts in the paths.  Gyan Reasoner takes the concept graphs of the concepts, finds all of their instances within the knowledge store and then finds all the paths between all the instances of each pair, while applying the constraints.

The $Synthesizer_{FindPath}$ in the simplest form can produce a Graph, or chain all the discourse units of the nodes in the paths as a listing or identify many more ways of synthesizing the information, depending on the verbosity settings.

\subsubsection{Aggregate}

The $Aggregate$ instruction is a powerful analytical task used to group a set of concepts based on a shared property and then perform a calculation (like counting) on those groups.  This usually requires working on top of other retrieval or analytical tasks to answer complex queries.  In some scenarios, this task is also available to the Synthesizer to create insightful data points to accompany the answers to simpler questions, depending on the Verbosity levels.

Essentially, an aggregate query can be implemented as a hierarchical query where the inner levels of the hierarchy provide retrieval instructions, and the outer levels of the query provide the aggregation instructions.  $INS=(AGGREGATE,[Query_1,Query_2,\ldots])$.  The $G_{Constraints}$ and $G_{Contexts}$ are typically empty in the outer Aggregate query but are fully or partially instantiated in the inner queries.

Gyan Reasoner recursively calls itself to gather all the instances of the inner queries.  Reasoner then performs various aggregations on the results and flags the ones with interesting characteristics.  The Reasoner then outputs the outcomes of these aggregations.

The ${Synthesizer}_{Aggregation}$ posses a number of options for producing the output.  It may provide all the descriptions of the interesting and uninteresting aggregations.  It may or may not provide details about the outcomes of the inner queries.  It may choose to present the descriptive statistics of the inner or outer results.  It may choose to include the detailed discourse units containing the outputs.  All of this can be handled through a Verbosity parameter.

\end{document}